\documentclass[letterpaper,journal]{IEEEtran}
\usepackage{amsmath,amsfonts}
\usepackage{array}
\usepackage[caption=false,font=normalsize,labelfont=sf,textfont=sf]{subfig}
\usepackage{textcomp}
\usepackage{stfloats}
\usepackage{url}
\usepackage{verbatim}
\usepackage{graphicx}
\usepackage{cite}
\usepackage{xcolor}    
\usepackage{colortbl}
\usepackage{arydshln}
\usepackage{tabularx}
\usepackage[ruled, lined, boxed, linesnumbered,  commentsnumbered, noend]{algorithm2e}
\usepackage{enumitem}
\usepackage{float}
\usepackage[colorlinks=true,linkcolor=blue,citecolor=blue]{hyperref}
\usepackage{arydshln}

\usepackage{booktabs}
\usepackage{enumitem}
\usepackage{wrapfig}
\usepackage{graphicx}
\usepackage{multirow}
\usepackage{makecell}

\usepackage{etoolbox} 
\definecolor{comment_color_2}{RGB}{64,128,128}
\newcommand{\LineComment}[1]{\small\textcolor{comment_color_2}{\textit{\# #1}}}

\definecolor{first}{HTML}{547CB1} %
\definecolor{improve}{HTML}{1E73C4} %
\newcommand{\firstone}[1]{\colorbox{first!25}{#1}}

\newcommand{\second}[1]{\colorbox{red!10}{#1}}

\newcommand{\upcolor}[1]{\textcolor{first}{#1}}

\newcommand{\myfont}{\fontfamily{ppl}\selectfont}
\newcommand{\mytext}[1]{\textbf{{\myfont #1}}}

\begin{document}

\title{MoA-VR: A Mixture-of-Agents System Towards All-in-One Video Restoration}

\author{Lu~Liu$^{*}$,~Chunlei~Cai$^{*}$,~Shaocheng~Shen,~Jianfeng~Liang,~Weimin~Ouyang,~Tianxiao~Ye,~Jian~Mao,

~Huiyu~Duan,~Jiangchao~Yao,~Xiaoyun~Zhang,~Qiang~Hu$^{\dagger}$,~\IEEEmembership{Member,~IEEE,}~Guangtao~Zhai,~\IEEEmembership{Fellow,~IEEE}
\thanks{Manuscript received May 31, 2025, revised September 30, 2025, accepted October 7, 2025. This work is supported by National Natural Science Foundation of China (62571322,  62271308, 62401365, 62225112, 62132006, U24A20220), STCSM ( 24ZR1432000, 24511106902, 24511106900, 22DZ2229005), 111 plan (BP0719010), China Postdoctoral Science Foundation (BX20250411, 2025M773473) and State Key Laboratory of UHD Video and Audio Production and Presentation. ($^{*}$ Equal contribution authors: Lu Liu and Chunlei Cai, $^{\dagger}$ Corresponding author: Qiang Hu)}

\thanks{Lu~Liu,~Shaocheng~Shen,~Huiyu~Duan,~Jianfeng Liang,~Jiangchao~Yao,~Xiaoyun~Zhang, ~Qiang~Hu and Guangtao Zhai are with the Shanghai Jiao Tong University, Shanghai, 200240, China. (email:{lettieliu, shenshaocheng, huiyuduan, blur\_damon, sunarker, xiaoyun.zhang, qiang.hu,  zhaiguangtao}@sjtu.edu.cn)

Weimin Ouyang is with University of Electronic Science and Technology of China, Hainan, 572400, China. (email:2022360903022@std.uestc.edu.cn)

Chunlei Cai, Tianxiao Ye and Jian Mao are with Bilibili Inc., Shanghai, 200433, China. (email:{caichunlei, yetianxiao, maojian}@bilibili.com)
}
}

\markboth{Journal of Selected Topics in Signal Processing,~Vol.~14, No.~30, May~2025}%
{Shell \MakeLowercase{\textit{et al.}}: A Sample Article Using IEEEtran.cls for IEEE Journals}


\maketitle

\begin{abstract}

Real-world videos often suffer from complex degradations, such as noise, compression artifacts, and low-light distortions, due to diverse acquisition and transmission conditions. Existing restoration methods typically require professional manual selection of specialized models or rely on monolithic architectures that fail to generalize across varying degradations. Inspired by expert experience, we propose MoA-VR, the first \underline{M}ixture-\underline{o}f-\underline{A}gents \underline{V}ideo \underline{R}estoration system that mimics the reasoning and processing procedures of human professionals through three coordinated agents: Degradation Identification, Routing and Restoration, and Restoration Quality Assessment. Specifically, we construct a large-scale and high-resolution video degradation recognition benchmark and build a vision-language model (VLM) driven degradation identifier. We further introduce a self-adaptive router powered by large language models (LLMs), which autonomously learns effective restoration strategies by observing tool usage patterns. To assess intermediate and final processed video quality, we construct the \underline{Res}tored \underline{V}ideo \underline{Q}uality (Res-VQ) dataset and design a dedicated VLM-based video quality assessment (VQA) model tailored for restoration tasks. Extensive experiments demonstrate that MoA-VR effectively handles diverse and compound degradations, consistently outperforming existing baselines in terms of both objective metrics and perceptual quality. These results highlight the potential of integrating multimodal intelligence and modular reasoning in general-purpose video restoration systems.

\end{abstract}

\section{Introduction}
\label{intro}
\begin{figure}
    \centering
    \includegraphics[width=1\linewidth]{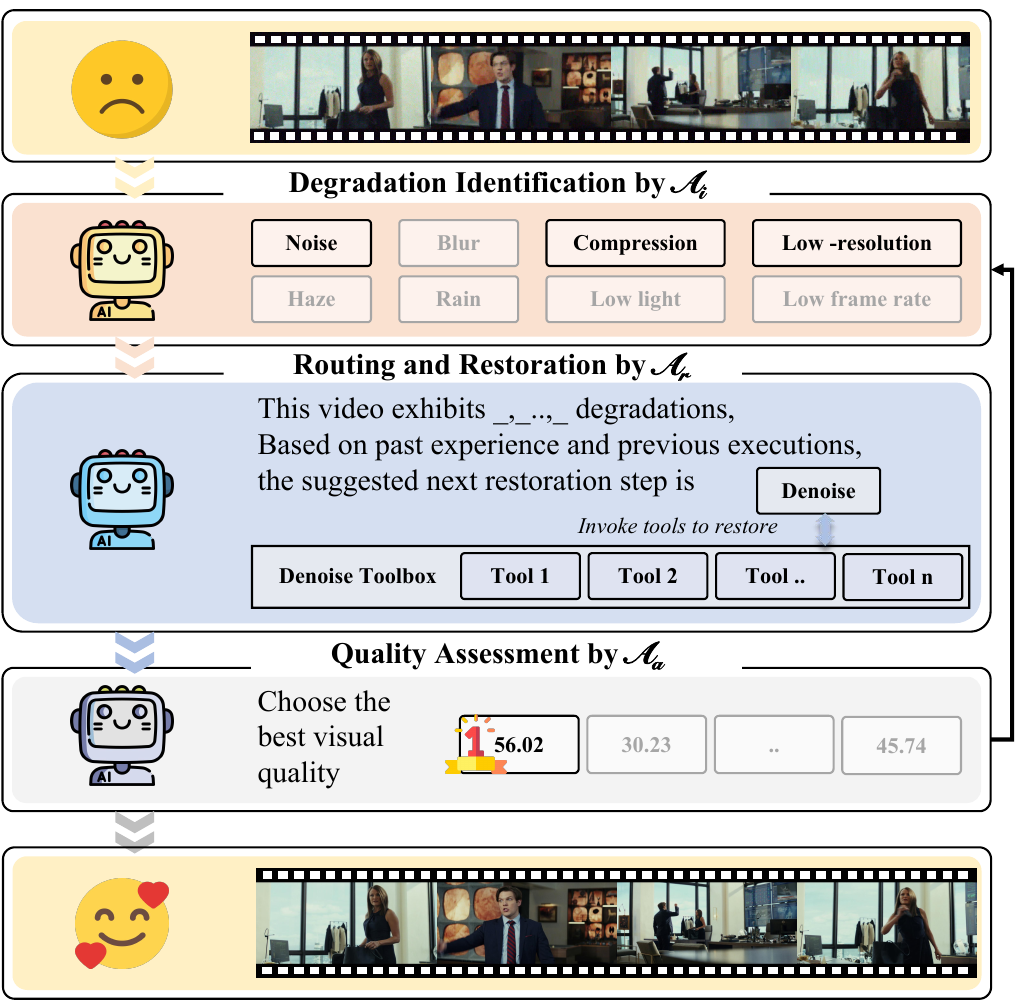}
    \caption{Overview of the agents in MoA-VR. MoA-VR restores low-quality video clips with complex degradations through the collaboration of three agents: the degradation identification agent, the routing and restoration agent, and the quality assessment agent.}
    \label{fig:1}
\end{figure}

The growing ubiquity of high-quality video data has fueled advances in diverse applications such as entertainment, surveillance, and autonomous driving, \textit{etc}. However, videos captured or transmitted in real-world environments often suffer from complex and heterogeneous degradations, including noise, compression artifacts, motion blur, and low-light conditions, \textit{etc}. These degradations, which may co-occur or vary spatially and temporally, significantly impair visual quality and downstream processing. Although video restoration (VR) techniques have seen notable progress, achieving robust and high-quality restoration in an all-in-one manner remains a formidable challenge due to the diversity, overlap and unpredictability of degradation types.


Traditional video restoration methods~\cite{wu2024rainmamba,xu2023map,Pan_2023_CVPR,li2023fastllve,guo2024generalizable,rota2024enhancing,chan2022investigating,zhou2024upscaleavideo} typically employ a multi-model combination strategy, where each specialized model is trained to handle a specific type of degradation. Representative works such as BasicVSR++~\cite{chan2022generalization} and VRT~\cite{liang2022vrt} have demonstrated promising results by effectively targeting particular degradation scenarios. However, it is very difficult for these specialized methods to achieve good performance in practical applications, as they heavily rely on accurately identifying the type of degradation, leveraging expert knowledge to carefully select appropriate models, and performing extensive visual assessment to choose the best result. Moreover, when multiple models need to be combined to handle complex real-world degradations, improper execution order can easily lead to error accumulation, resulting in poor robustness and unsatisfactory performance.



In contrast, all-in-one models~\cite{liang2022rvrt,NEURIPS2024_e635a25e,liang2022vrt,chan2022generalization} aim to streamline the restoration process by using a single unified network to handle diverse degradation types. While recent advances such as AverNet~\cite{NEURIPS2024_e635a25e} demonstrate promising generalization across tasks, these models often rely on increased capacity and are limited by their static architecture. In practical deployments, especially in large-scale or resource-constrained video applications, model size constraints, domain variability, and unknown degradation compositions make it difficult to achieve consistent performance. As a result, even all-in-one solutions are typically deployed as multiple specialized smaller distilled models. Proper use of  them still requires manual tuning and massive evaluation. This reveals a key limitation: static restoration pipelines struggle with real-world complexity, highlighting the necessity for dynamic and adaptive solutions.

Recently, large language models (LLMs), such as DeepSeek~\cite{guo2025deepseek}, and vision-language models (VLMs), such as BLIP~\cite{li2022blip}, GPT-4o~\cite{GPT4}, and LLaVA~\cite{li2024llava}, have shown remarkable success in multimodal understanding and reasoning tasks. These models inherently possess rich prior knowledge and strong contextual reasoning capabilities, making them well-suited for interpreting complex visual information alongside natural language descriptions. Such abilities suggest a promising direction for video restoration, where diverse and compound degradations need to be accurately identified and adaptively addressed. Since traditional methods struggle with manual degradation classification and fixed restoration pipelines, and all-in-one models lack flexibility and fine-grained control, by leveraging the adaptive reasoning power of VLMs and LLMs, it is possible to build more intelligent and autonomous video restoration systems that dynamically understand and respond to the unique degradation characteristics of each video. However, the application of VLMs and LLMs in video restoration remains largely underexplored, particularly in tasks requiring precise degradation identification and adaptive restoration strategy selection.

To bridge this gap, we propose MoA-VR, a novel intelligent video restoration framework that integrates multimodal perception and modular reasoning within a unified framework, referring the cognitive processes of human experts. MoA-VR decomposes the restoration process into three fundamental capabilities, Identification, Routing, and Assessment, mimicking how human professionals analyze visual degradations, plan restoration workflows, and judge output quality as show in Fig.\ref{fig:1}. By introducing a modular architecture empowered by VLMs and LLMs, MoA-VR brings flexibility and adaptability to the restoration of complex, mixed, and temporally non-uniform degradations across video sequences. This design enables our system to make task-aware decisions and perform context-sensitive tool selection with minimal human supervision.

To accurately identify degradation types, we establish a vision-language model-based benchmark to systematically evaluate various VLMs on multi-type degradation perception, allowing the model to capture fine-grained and temporally variant artifacts such as compression noise, motion blur, and resolution drop. Based on the recognized degradation profiles, we introduce a self-adaptive routing module powered by LLMs, which learns to synthesize optimal restoration pipelines by interpreting tool execution traces and planning context-aware toolchains. To close the loop, a dedicated video quality assessment (VQA) module is developed and trained on the newly curated \underline{Res}toration \underline{V}ideo \underline{Q}uality (Res-VQ) dataset, which includes high-resolution video pairs and human-annotated perceptual quality scores. This enables MoA-VR to judge the restoration results and guide iterative optimization when needed. Extensive experiments demonstrate that MoA-VR significantly outperforms state-of-the-art video restoration methods, achieving a \textbf{3.02 dB} improvement in PSNR and notable gains in both perceptual and pixel metrics, while requiring minimal human intervention.

Our main contributions are summarized as follows:

\begin{itemize}
  \item We propose MoA-VR, the first multi-agent modular framework for intelligent all-in-one video restoration, integrating identification, routing, and assessment.
  \item We establish the first VLM-based benchmark for video degradation recognition, enabling accurate analysis of eight common real-world distortions.
  \item We design a self-adaptive routing module guided by LLMs that learns optimal restoration pipelines from historical tool usage traces.
  \item We curate the Restored Video Quality (Res-VQ) dataset with human-rated quality scores and build a dedicated VQA model tailored for restoration quality evaluation.
\end{itemize}

\begin{figure*}[h]
    \centering
    \includegraphics[width=1\linewidth]{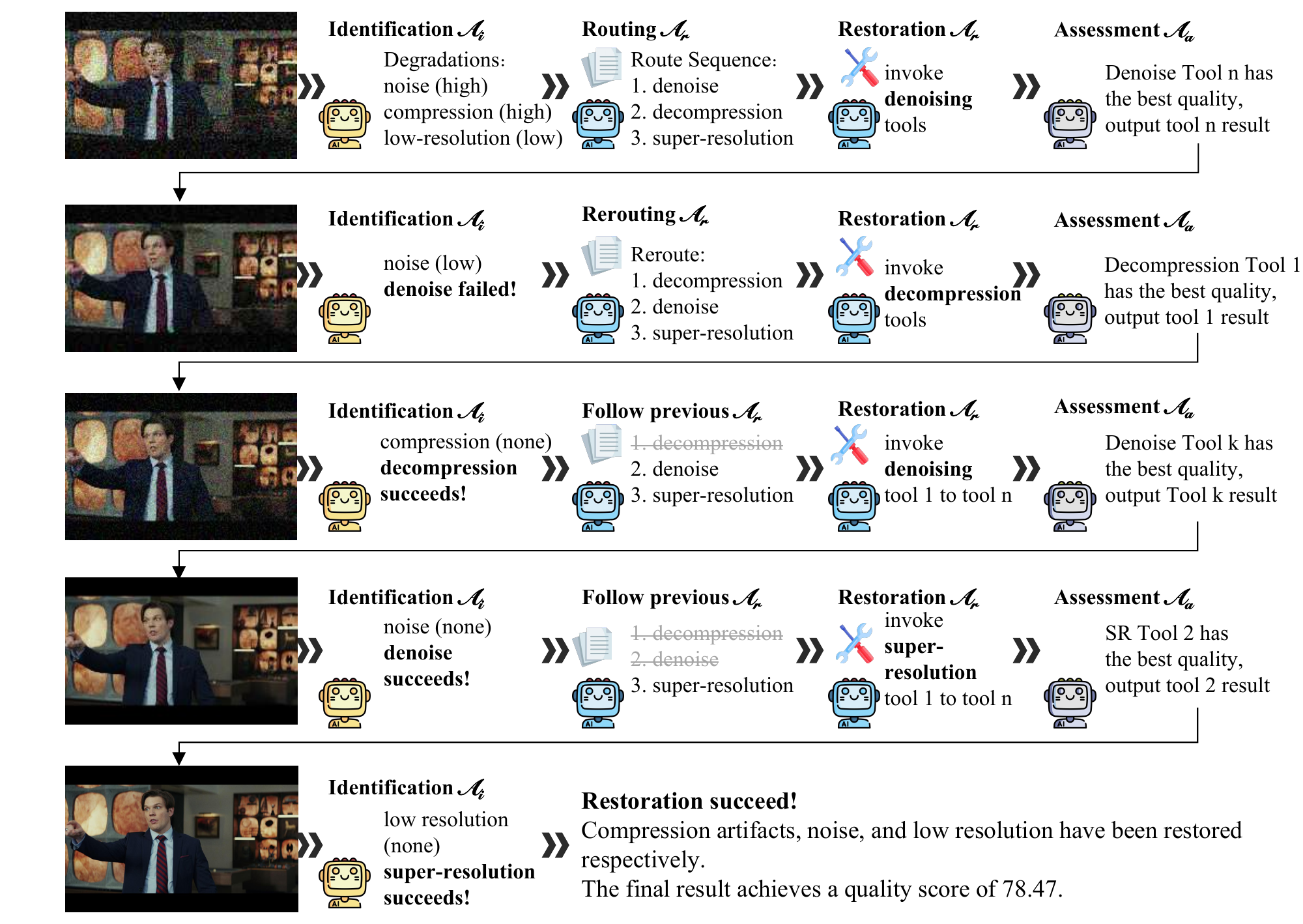}
    \caption{ MoA-VR incorporates three specialized agents within a closed-loop architecture. For a low-quality input video, $\mathcal{A}_i$ identifies the degradation type and level; $\mathcal{A}_r$ generates a degradation removal plan and then invokes the corresponding restoration toolbox; $\mathcal{A}_a$ assesses all the intermediate results and chooses the best quality one. Then $\mathcal{A}_i$ identifies whether the previous restoration was successful. If it fails, $\mathcal{A}_r$ rolls back and reroutes; if successful, $\mathcal{A}_r$ follows the previous plan. This loop continues until all degradations are removed.}
    \label{fig:workflow}
\end{figure*}

\section{Related Works}
\label{rw}

\subsection{Video Restoration}

Video restoration is a fundamental research field focused on reconstructing high-quality video content from degraded observations. Beyond the application of single-frame image methodologies~\cite{duan2024uniprocessor,9769950,gao2024qualityguidedskintoneenhancement}, current VR methods can be categorized into two paradigms based on their application scenarios: task-specific degradation processing models~\cite{wu2024rainmamba,xu2023map,Pan_2023_CVPR,li2023fastllve,guo2024generalizable,varfvv} and all-in-one models~\cite{liang2022rvrt,NEURIPS2024_e635a25e,liang2022vrt,chan2022generalization}. Task-specific degradation processing models are designed to address individual degradation types through dedicated learning frameworks. These models typically require prior human expert assessment to identify the video degradation characteristics before processing. 


However, real-world video degradation typically involves complex mixed artifacts, making single-task restoration methods fundamentally limited. To address this, VRT~\cite{liang2022vrt} and BasicVSR++~\cite{chan2022generalization} proposed a unified framework trained simultaneously across multiple tasks, including denoising, deblurring, super-resolution and so on. Moreover, AverNet~\cite{NEURIPS2024_e635a25e} effectively handle previously unseen degradation patterns by demonstrating a single set of weights. Although all-in-one video restoration models have demonstrated preliminary success, their robustness in handling diverse unknown degradations remains insufficient, often requiring manual post-processing by human experts. This critical limitation motivates the urgent need for an intelligent framework capable of automatically identifying degradation patterns and performing adaptive restoration.




\subsection{LLMs and VLMs}

Large Language Models~\cite{guo2025deepseek,openai2024gpt4technicalreport,meta2024llama} and Vision-Language Models~\cite{chen2024expanding,chen2024expanding,wang2024mpo} have emerged as pioneering forces in advancing general-purpose artificial intelligence. Trained on massive-scale multimodal datasets encompassing both textual and visual data, these models demonstrate remarkable problem-solving capabilities that extend far beyond conventional language processing tasks. Some studies extend LLMs to specialized domains, such as mathematical problem-solving and legal query processing~\cite{lawgpt}, further blurring the line between human and machine intelligence. Meanwhile, models such as GPT-4o~\cite{GPT4} and Qwen-Serials~\cite{Qwen2.5-VL} demonstrate unprecedented performance in real-world multimodal interactions. The demonstrated cognitive and perceptual competencies of LLMs and VLMs have established them as foundational architectures for developing sophisticated multi-agent decision systems.


\subsection{Multi-Agent System}
Multi-agent systems (MAS) consist of multiple interacting intelligent agents that collaboratively or competitively solve complex problems~\cite{1995Intelligent,hong2025dual,cao2025analytical}.  Early MAS research primarily relied on symbolic reasoning~\cite{shoham2008multiagent} and game-theoretic approaches~\cite{10.5555/3091574.3091594}, which provided strong theoretical guarantees but struggled with scalability in open-world environments. Recent advances in foundation models have fundamentally transformed this paradigm. Specifically, the emergence of LLMs and VLMs has introduced new paradigms for multi-agent coordination. LLMs enable agents to engage in natural language communication, facilitating role allocation, negotiation, and strategy formulation. For instance, Generative agents~\cite{10.1145/3586183.3606763} demonstrated that LLM-driven agents can simulate human-like social behaviors in virtual environments, while HuggingGPT~\cite{NEURIPS2023_77c33e6a} orchestrates multiple AI models through LLM-based task decomposition.

Moreover, VLMs further extend multi-agent capabilities into visually grounded environments. In virtual environments, Ghost in the Minecraft~\cite{zhu2023ghost} demonstrated hierarchical planning for long-horizon tasks through visual-language grounding. For distributed perception, AVLEN~\cite{li2023generalist} developed attention mechanisms for VLM-equipped drone swarms, enabling emergent coordination during exploration. In the domain of image restoration research, AgenticIR~\cite{agenticir} and Restoreagent~\cite{chen2024restoreagent} independently developed novel multi-agent frameworks incorporating VLMs, marking a significant breakthrough by extending multi-agent system applications to image inpainting tasks.

\begin{table*}[ht]
  \centering
  \caption{Comparison of popular video restoration datasets. This table includes datasets for various tasks such as super-resolution, deblurring, and compression artifact removal.}
  \vspace{-10pt}
  \renewcommand\arraystretch{1}
  \resizebox{\textwidth}{!}{
    \begin{tabular}{lccccc}
      \toprule[1pt]
      \bf Dataset & \bf Task Type & \bf Resolution & \bf Clips & \bf Total Frames & \bf Frames per Clip  \\
      \midrule
      GoPro~\cite{Nah_2017_CVPR}& Deblur & 1280×720 & 33 & 3214 & 97  \\
      DVD~\cite{dvd} & Deblur & 1280×720 & 71 & 5708 & 95  \\
      BSD~\cite{bsd} & Deblur & 640×480 & 300 & 30000 & 100  \\
      REDS~\cite{reds} & SR, Deblur & 1280×720 & 300 & 30000 & 100  \\
      Vimeo-90K~\cite{vimeo90k} & SR, Denoising, Decomp. & 448×256 & 91,701 & 641,970 & 7  \\
      \hdashline
      \rowcolor{gray!20} 
       MoA-VD& 
      \makecell[c]{8 tasks: SR, Denoising, Decompression, Deblur, \\Low-light Enhancement, Dehazing, Deraining, Interpolation} & 
      1920×1080 & 3,300 & 330,000 & 100  \\
      \bottomrule[1pt]
    \end{tabular}
  }
  \label{tab:video_datasets}
\end{table*}

\subsection{Visual Quality Assessment}
For Visual Quality Assessment, conventional approaches~\cite{quality:BMPRI,tu2021ugc,tu2021rapique,Duan_2022} typically evaluate video quality through handcrafted feature extraction and regression analysis. With the advancement of deep learning techniques, numerous deep neural network (DNN)-based VQA models have been proposed and demonstrated superior performance. Several representative approaches~\cite{sun2022deep,wu2022fast,wu2023exploring,chen2021learning,duan2023attentive,duan2022develop,yang2025omni} employ pre-trained deep neural networks to extract semantic features and predict quality scores through training regression evaluator. However, these methods exhibit limited capability in restoration-type understanding and can only provide holistic quality assessments for videos.

\section{Methodology}
\label{method}

\subsection{Problem Formulation and System Architecture} 
\label{sec3.1}
We consider a comprehensive degradation space \(\mathcal{D} = \{d_1, d_2, \ldots, d_n\}\), where each \(d_i\) denotes a specific degradation type, such as noise, blur, compression artifacts, rain, haze, low resolution, low frame rate, or low-light conditions. For each degradation \(d_i\), we maintain a dedicated toolset \(\mathcal{T}_{d_i} = \{T_{d_i}^1, T_{d_i}^2, \ldots\}\), where each tool \(T_{d_i}^j\) is tailored to address \(d_i\) under varying contexts and severity levels.
Given a degraded video \(V\) potentially affected by an unknown combination of degradations from \(\mathcal{D}\), our objective is to design an agentic system that can iteratively identify active degradations, determine an effective removal sequence, select appropriate tools, and refine its strategy based on restoration quality feedback. Formally, the restoration process is defined as:
\[
\tilde{V} = \mathcal{F}(V),
\]
where \(\tilde{V}\) denotes the restored video output.

To this end, we propose MoA-VR, a modular multi-agent video restoration system inspired by the collaborative workflow of human video repair experts. As illustrated in Fig.~\ref{fig:workflow}, MoA-VR comprises three VLM-empowered agents, each specializing in a critical aspect of video restoration: the \textit{Degradation Identification Agent} \(\mathcal{A}_i\), the \textit{Routing and Restoration Agent} \(\mathcal{A}_r\), and the \textit{Quality Assessment Agent} \(\mathcal{A}_a\).

The Degradation Identification Agent \(\mathcal{A}_i\) performs comprehensive analysis of the input video, detecting degradation types and their severity levels (e.g., low, medium, high). It generates a structured diagnostic output that serves as a precise foundation for restoration planning. This agent is implemented via a fine-tuned vision-language model, enabling robust and context-aware degradation diagnosis.

Based on the diagnostic output, the Routing and Restoration Agent \(\mathcal{A}_r\) formulates an explicit restoration sequence, decomposing the task into subtasks targeting specific degradations. It then selects and applies restoration tools from the respective toolsets \(\mathcal{T}_{d_i}\). Importantly, \(\mathcal{A}_r\) iteratively refines the restoration plan by incorporating feedback from intermediate restoration results, enabling a flexible, adaptive, and content-aware enhancement process.

To ensure high restoration quality, the Quality Assessment Agent \(\mathcal{A}_a\) acts as an automated evaluator, estimating the visual quality of intermediate outputs. By assigning quantitative quality scores, it assists \(\mathcal{A}_r\) in selecting the most effective tools for each subtask. This agent emulates the human-in-the-loop quality inspection process, thereby enhancing robustness and decision reliability.

Together, these three agents form a tightly coupled closed-loop system that iteratively processes the input video: \(\mathcal{A}_i\) diagnoses degradations, \(\mathcal{A}_r\) executes and adapts restoration operations, and \(\mathcal{A}_a\) provides continuous quality feedback. This collaborative architecture offers strong flexibility, scalability, and generalization capacity, effectively supporting robust restoration across diverse and complex degradation scenarios.


\subsection{Degradation Identification Agent \(\mathcal{A}_i\)}
\label{sec3.2}
\begin{figure}
    \centering
    \includegraphics[width=1\linewidth]{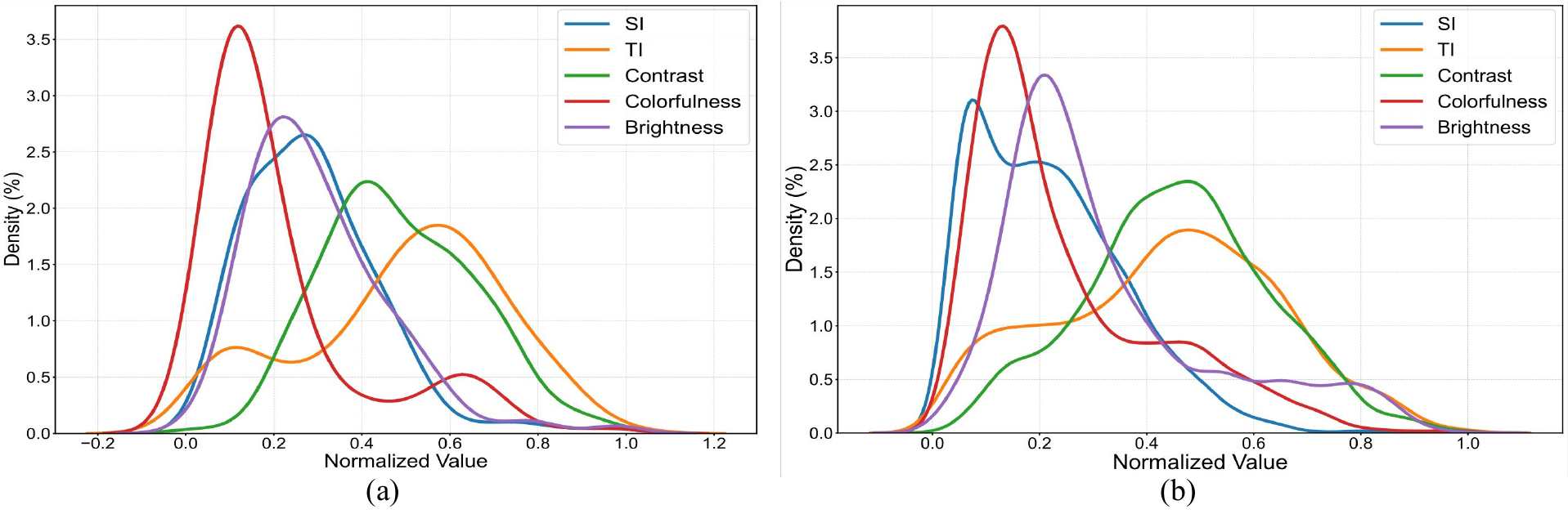}
    \caption{Feature distribution of (a) MoA-VD-GT and (b) MoA-VD-LQ. SI and TI indicate spatial and temporal information, respectively.}
    \label{fig:3.1}
\end{figure}

To effectively support downstream restoration agents, the Degradation Identification Agent \(\mathcal{A}_i\) is introduced to recognize both the type and severity of degradations present in input videos. This subsection elaborates on the construction of a comprehensive training dataset and the multi-modal model design of \(\mathcal{A}_i\), highlighting its capabilities in diverse and compound degradation recognition.

\textbf{Dataset Construction.} Training a robust degradation identifier demands a large-scale, high-fidelity dataset with fine-grained annotations. Existing video restoration datasets, summarized in Table~\ref{tab:video_datasets}, often suffer from limited diversity in degradation types and insufficient resolution. To overcome these limitations, we construct the MoA-VD dataset, featuring 300 high-quality 1080P-resolution videos paired with 3000 degraded versions. The source content consists of 300 diverse 4K videos, including humans, objects, scenes, and animations, collected from online platforms and downsampled to 1080P using bicubic interpolation to preserve quality while avoiding upsampling artifacts. The feature diversity of MoA-VD is shown in Fig.~\ref{fig:3.1}, covering a wide range of color distributions, contrast variations, and temporal dynamics.

To generate degraded counterparts, we design a comprehensive degradation pipeline inspired by Real-ESRGAN~\cite{wang2021realesrgan} and AgenticIR~\cite{agenticir}, simulating eight common real-world degradation types: rain, haze, blur, low resolution, low frame rate, low light, noise, and compression artifacts. Each degraded video is labeled with its degradation type(s) and corresponding severity levels (low, medium, high). Notably, both single and mixed degradations are included, enhancing model generalizability.

\begin{figure}
    \centering
    \includegraphics[width=1\linewidth]{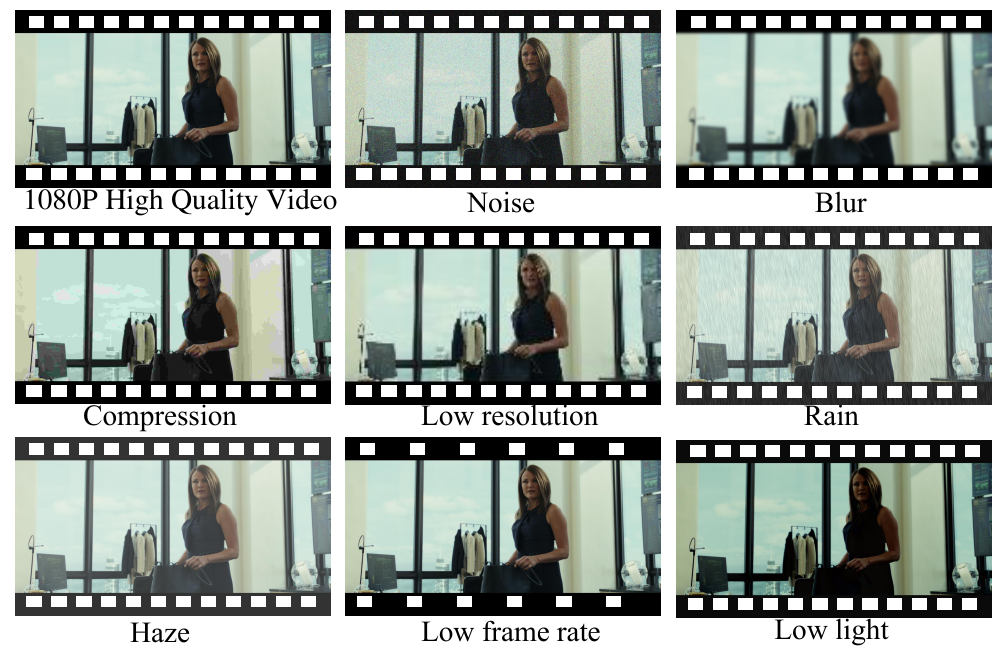}
    \caption{Visual Examples of Different Video Degradations}
    \label{fig:3.2new}
\end{figure}

Specifically, haze is generated using an atmospheric scattering model with depth-based transmission estimation; rain is simulated via directional filtering and Gaussian noise perturbations; blur includes both defocus (circular kernel) and gaussian blur; noise covers Gaussian and Poisson variants with adjustable intensity; low-light degradation is achieved by modifying the luminance channel in HSV color space; low resolution is obtained through bicubic downsampling; low frame rate via frame dropping; and compression artifacts are introduced by H.264 encoding. For mixed degradations, up to three types are randomly selected and applied sequentially in a realistic order. The resulting dataset consists of 300 ground-truth videos and 3000 degraded clips, totaling 300,000 frames, as illustrated in Fig.~\ref{fig:3.2new}.

\textbf{Degradation Identification.} The Degradation Identification Agent $\mathcal{A}_i$ is designed as a VLM capable of multi-modal reasoning to identify degradation types and assess their severity in video content. Inspired by recent advances in instruction-tuned multimodal learning~\cite{duan2024finevq,f-bench,xu2025lmm4edit}, we reformulate the degradation assessment task as a \textit{text-based classification problem}, enabling the model to better exploit the semantic priors embedded in large language models  and improve interpretability.

\textcolor{black}{The degradation identification agent is capable of identifying eight degradation types: noise, blur, compression, low resolution, rain, haze, low frame rate, and low light. These types were selected based on their frequency in real-world scenarios, where degradations often appear as combinations of these factors. In comparison to other all-in-one restoration methods that typically address only a limited set of degradations, our approach considers a broader range, as supported by prior works~\cite{duan2024uniprocessor, AirNet} and real-world applications. These degradations are among the most common and urgent to address in all-in-one restoration tasks. Some other less typical degradations may be considered for future work.}
 
\begin{figure}
    \centering
    \includegraphics[width=1\linewidth]{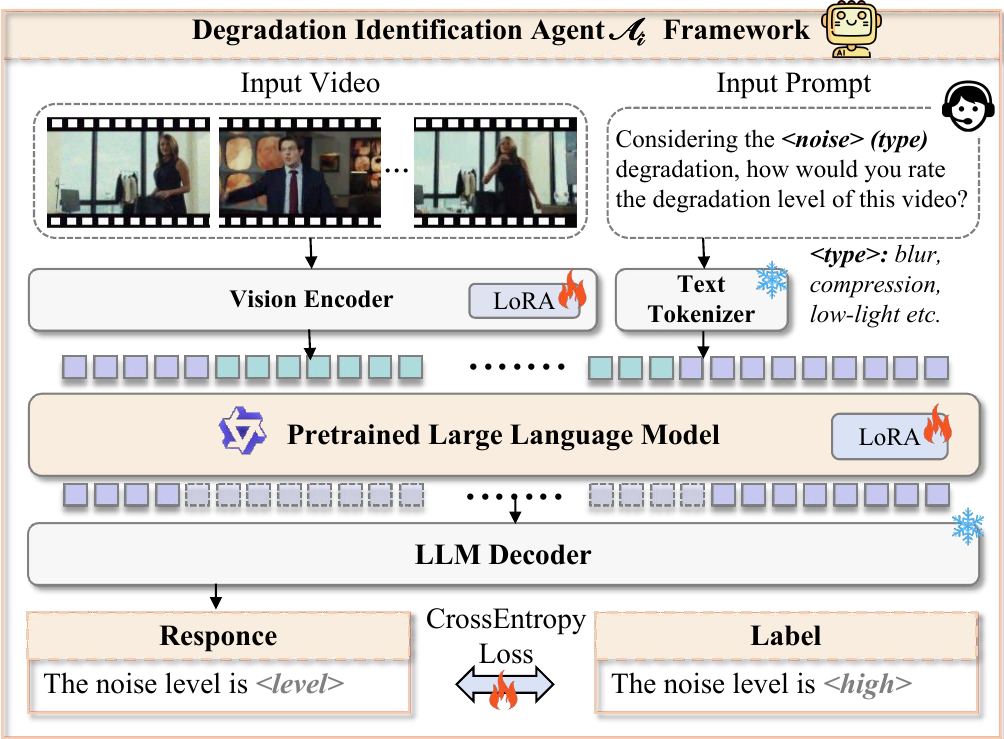}
    \caption{The overall framework of \(\mathcal{A}_i\). \(\mathcal{A}_i\) can evaluate all types of degradation levels in an all-in-one framework. It can process videos, along with prompts, to identify the degradations. It consists of a vision encoder to extract both spatial and temporal features and a text tokenizer to tokenize the input prompts. These features are projected into the same space by trained projectors. A pre-trained LLM is utilized to fuse the features while fine-tuned with LoRA.}
    \label{fig:3}
\end{figure}

As depicted in Fig.~\ref{fig:3}, the input video is first processed by the vision encoder Qwen-VL-2.5-ViT~\cite{Qwen2.5-VL}, which has been fine-tuned using Low-Rank Adaptation (LoRA) to capture rich spatial-temporal representations efficiently. The extracted visual features are then projected into the language embedding space via two stacked multilayer perceptrons (MLPs). These projected visual embeddings are concatenated with tokenized textual prompts to form joint multi-modal tokens. Subsequently, these fused representations are passed into the pre-trained language backbone Qwen-VL2.5-7B~\cite{Qwen2.5-VL} for contextual understanding and multi-modal inference.

Rather than regressing scalar values for degradation levels, $\mathcal{A}_i$ generates human-readable textual statements, such as \textit{``The noise level is medium.''} This formulation aligns more naturally with the strengths of LLMs, which are more adept at modeling discrete semantic categories than continuous numerical values. Moreover, this approach enhances interpretability and generalizability to unseen degradation types or ambiguous cases.

To adapt the pre-trained VLM to the degradation classification task, we apply LoRA~\cite{hu2021lora} to both the vision encoder and the LLM. Specifically, for a frozen linear transformation $h = Wx$, LoRA introduces a low-rank update:
\begin{equation}
h = Wx + \Delta Wx = Wx + \frac{\alpha}{r} BAx,
\end{equation}
where $A \in \mathbb{R}^{r \times d_i}$, $B \in \mathbb{R}^{d_o \times r}$, and $r \ll \min(d_o, d_i)$ is the rank of the adaptation. The scalar $\alpha$ is a learnable scaling factor. 

\textcolor{black}{We fine-tuned the LLM and vision encoder by minimizing a token-level cross-entropy loss. Formally, the language loss is defined as:
\begin{equation}
\mathcal{L}_{\text{language}} = -\frac{1}{N} \sum_{i=1}^N \log P(y_{\text{label}} \mid y_{\text{pred}}),
\end{equation}
where $y_{\text{pred}}$ is the predicted token, $y_{\text{label}}$ is the ground truth token, $P(y_{\text{label}} \mid y_{\text{pred}})$ is the predicted probability of the correct token, and $N$ is the total number of tokens. This fine-tuning strategy ensures lightweight yet effective adaptation, allowing $\mathcal{A}_i$ to deliver accurate, fine-grained classification of degradation types and severities. The identified degradation cues provide essential semantic guidance for downstream restoration agents.
}

\subsection{Routing and Restoration Agent \(\mathcal{A}_r\)}
\label{sec3.3}

To tackle the diversity and context-dependency inherent in video degradations, we propose the Routing and Restoration Agent \(\mathcal{A}_r\) that adaptively plans multi-step degradation removal sequences by leveraging LLM-guided reasoning combined with a self-adapted mechanism. This agent orchestrates the restoration process by integrating high-level decision-making with modular, degradation-specific restoration tools.

Inspired by recent advances in agent-based image restoration~\cite{agenticir, restoreagent}, we extend adaptive degradation routing into the more complex video domain, where the space of possible degradations and their combinations grows combinatorially. Static or rule-based routing methods quickly become insufficient to address this challenge. To enable scalable and intelligent routing, \(\mathcal{A}_r\) employs GPT-4~\cite{openai2024gpt4technicalreport} as a high-level reasoning engine. For each degraded input, \(\mathcal{A}_r\) prompts GPT-4 to generate candidate sequences of degradation removal steps. Although GPT-4 demonstrates strong abstract reasoning capabilities~\cite{wei2023chainofthoughtpromptingelicitsreasoning}, it initially lacks specific expertise in video restoration, which can lead to suboptimal routing proposals.

To bridge this gap, we introduce a self-exploration framework that allows \(\mathcal{A}_r\) to iteratively improve its routing strategy. The agent experiments with multiple degradation removal sequences for each input, evaluating restoration quality with support from the Degradation Identification Agent \(\mathcal{A}_i\). Each trial, whether successful or not, is recorded as experience. Periodically, GPT-4 consolidates these accumulated experiences into a task-specific routing knowledge base, refining its future planning to better handle composite and previously unseen degradation patterns.

\begin{figure}[tb]
    \centering
    \includegraphics[width=1\linewidth]{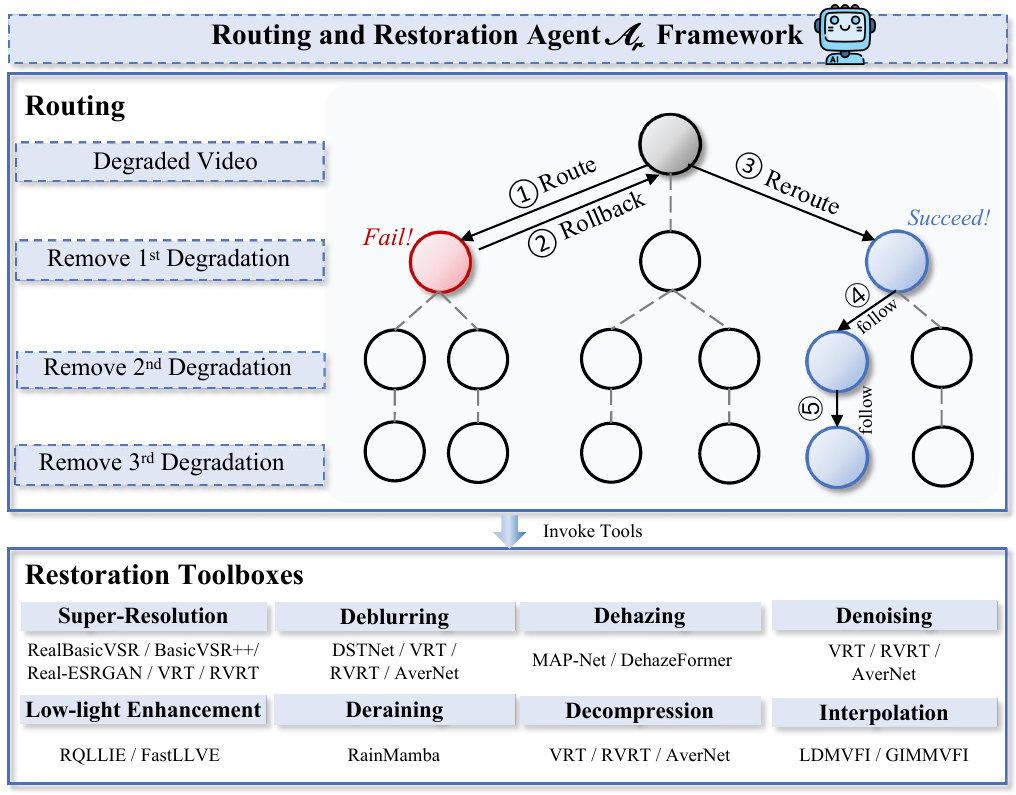}
    \caption{Illumination of degradation removal process by Routing and Restoration Agent \(\mathcal{A}_r\).  \(\mathcal{A}_r\) is able to route the degradation removal orders, rollback when restoration fails, reroute to another degradation removal orders. }
    \label{fig:route}
\end{figure}

Given the inherent complexity of multi-step restoration, failures in routing can still occur. To enhance robustness, \(\mathcal{A}_r\) incorporates a rollback and rerouting mechanism. Upon encountering ineffective degradation removal, the agent backtracks to a prior state, excludes the failed route, and generates an alternative plan. Failed sequences are cached to avoid repetition and inform subsequent explorations (see Fig.~\ref{fig:route}).

The actual restoration steps are executed by invoking a suite of modular restoration toolboxes, each specialized for a particular degradation type such as blur, noise, or compression artifacts. These toolboxes integrate state-of-the-art open-source models and can be updated flexibly as new methods emerge. Once \(\mathcal{A}_r\) selects a degradation class, the corresponding toolbox is applied sequentially to the degraded input. We select 1 to 4 tools for each task. For example, BasicVSR++~\cite{chan2022generalization} for video super-resolution and VRT ~\cite{liang2022vrt} for deblurring.

By synergizing LLM-guided strategic planning, iterative experience-driven refinement, failure-aware rollback, and modular restoration toolbox, \(\mathcal{A}_r\) achieves highly adaptive, scalable, and robust restoration performance across a broad spectrum of video degradations.

\subsection{Quality Assessment Agent \(\mathcal{A}_a\)}
\label{sec3.4}
\begin{figure*}[h]
    \centering
    \includegraphics[width=1\linewidth]{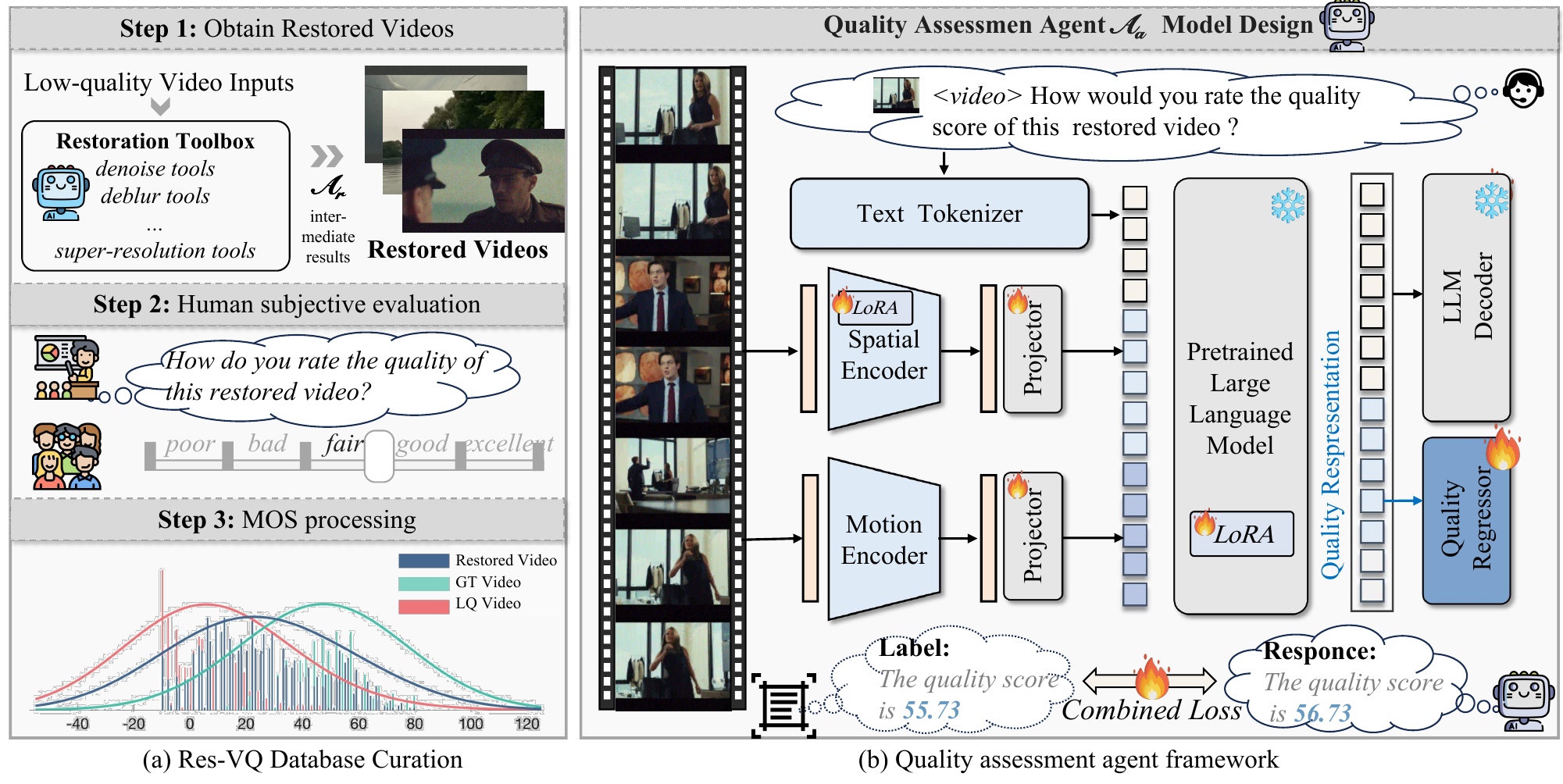}
    \caption{\textcolor{black}{(a) Res-VQ quality assessment dataset curation process. (b) An overview of quality assessment agent. It consists of three feature encoders, including an image feature extractor for extracting spatial features from sparse video frames, a motion feature extractor for extracting motion features from the entire video, and a text encoder for extracting aligned text features from prompts. The extracted features are then aligned through projectors and fed into a pre-trained LLM to generate the output results. LoRA weights are introduced to the pre-trained image encoder and the large language model to adapt the models to the quality assessment task.}}
    \label{fig:enter-label}
\end{figure*}

\textcolor{black}{In our agent-based restoration system, reliable video quality assessment (VQA) is crucial for guiding the self-evolution of routing policies and evaluating restoration effectiveness in alignment with human preferences. Unlike traditional User-generated Content (UGC) video quality assessments, our VQA evaluates restored videos generated by our system, which exhibit distinct characteristics. However, directly applying VQA models trained on UGC videos often fails to accurately reflect human preferences, and there is currently no dedicated database for restored videos quality assessment. To address this gap, we develop the Res-VQ dataset, specifically annotated for restored videos and capturing the features introduced during the restoration process. Based on this dataset, we propose a tailored VQA agent that better evaluates restored content and aligns more closely with human preferences.}

\textbf{Restored Video Dataset Construction and Subjective Scoring.}
\textcolor{black}{To train a quality assessment model capable of robustly evaluating diverse restoration outcomes, we first construct Res-VQ, a subjective quality dataset comprising 2,000 restored video outputs generated by our agentic system. Res-VQ contains both the restored clips and their corresponding human mean opinion scores (MOSs), and is used exclusively for training and evaluating the quality assessment agent. Compared with the previously proposed MoA-VD, these two datasets are disjoint and serve orthogonal purposes: MoA-VD focuses on the restoration process and consists of LQ–HQ video counterparts with degradation labels, whereas Res-VQ is dedicated to perceptual quality prediction of restored outputs.}

\textcolor{black}{For source video collection, we randomly sample 1,500 intermediate outputs from our system across different stages of degradation removal, meaning that these clips have been restored by the restoration tools one or more times. To ensure quality diversity, we additionally include 250 original low-quality inputs and 250 corresponding high-quality video references in the source video set. These source videos span a wide range of degradation types, restoration methods, and restoration plan sequences, ensuring diversity and realism. }

For ground-truth annotation, we conduct a controlled subjective study with 15 human raters having normal or corrected-to-normal vision. Prior to rating, all participants undergo expert-led training to calibrate their perception of restoration quality. During the experiment, videos are displayed in randomized order on 27-inch 4K monitors under standardized lighting conditions. Each clip is rated on a 5-point scale based on perceived restoration effectiveness. The protocol strictly follows ITU-R BT.500-14~\cite{series2012methodology} guidelines to ensure consistency and validity.

To process subjective scores, we apply kurtosis-based outlier rejection with a 3\% threshold to reduce inter-rater variance. The remaining scores are normalized into a [0,100] range using z-score standardization and linear rescaling:
\begin{eqnarray}
\begin{aligned}
z_i{}_j=\frac{r_i{}_j-\mu_j}{\sigma_i}, ~\quad z_{ij}'=\frac{100(z_{ij}+3)}{6},
\end{aligned}
\end{eqnarray}
\begin{eqnarray}
\begin{aligned}
\mu_j=\frac{1}{M_i}\sum_{i=1}^{M_i}r_i{}_j, ~\sigma_i=\sqrt{\frac{1}{M_i-1}\sum_{j=1}^{M_j}{(r_i{}_j-\mu_j)^2}},
\end{aligned}
\end{eqnarray}
where $r_{ij}$ denotes the score from the $i$-th participant on video $j$, and $M_i$ is the number of rated samples for participant $i$. Final mean opinion scores (MOS) are calculated by averaging the normalized scores:
\begin{eqnarray}
\begin{aligned}
MOS_j=\frac{1}{N}\sum_{i=1}^{N}z_{ij}'
\end{aligned}
\end{eqnarray}
where $N$ is the number of valid participants, and $z'_i{}_j$ denotes rescaled z-scores.

\textbf{Vision-Language Quality Regression Model.}
\textcolor{black}{To model quality in a way that aligns with human judgment while scaling across unseen degradations, we build a VLM-based quality assessor. Compared to classification, regression tasks impose stricter requirements on the fidelity of visual representations, especially for subtle perceptual differences in restored outputs. Our model incorporates both spatial and temporal cues: visual frames are processed by a vision encoder, and motion information is captured using the SlowFast network~\cite{slowfast}. These representations are projected into a shared token space and concatenated. However, as prior studies indicate that VLMs often struggle with numerical precision, we introduce a lightweight two-layer multilayer perceptron as a quality regressor to decode final scores. The hidden state from the token immediately preceding the numeric prediction in the LLM sequence is extracted and passed to a quality decoder. 
We optimize the training using a combination of language loss and L1 loss. 
The training objective combines language loss and L1 loss.
The L1 loss is defined as:
\begin{equation}
    \mathcal{L}_1 = \frac{1}{N} \sum_{i=1}^{N} |q_\text{pred} - q_\text{label}|,
\end{equation}
where $q_\text{pred}$ and $q_\text{label}$ are the predicted and ground truth quality scores, and $N$ is the number of videos in a batch.  
The overall loss function is the sum of the two:
\begin{equation}
    \mathcal{L}_{\text{combined}} = \mathcal{L}_{\text{language}} + \mathcal{L}_1.
\end{equation}
This architecture benefits from the interpretability and generalization of LLMs while addressing their inherent limitations on quantitative prediction tasks. }

\subsection{Agent Collaboration and Closed-Loop Design}
\label{sec3.5}

Building upon the specialized capabilities of the three agents introduced earlier, we now detail their collaborative interaction within a unified closed-loop framework designed for robust and adaptive video restoration. This framework emulates human-like iterative problem-solving by leveraging inter-agent communication, feedback-driven decision making, and dynamic adjustments to progressively remove complex and mixed degradations. An overview of the MoA-VR workflow is illustrated in Fig.~\ref{fig:workflow}, and the corresponding agent collaboration process is detailed in Algorithm~\ref{alg:unified_restoration_workflow}.


Given an input video suffering from unknown and compound degradations, the restoration process begins with the Degradation Identification Agent $\mathcal{A}_i$, which performs a fine-grained analysis to detect the presence and severity of each degradation type. This is achieved by querying all known degradation categories and applying a predefined \textit{low-level} threshold as the detection criterion. The resulting degradation profile serves as the initial condition for subsequent routing decisions.

\begin{table*}[ht]
\caption{
Comparison of MoA-VR with All-in-One methods for multi-degraded video restoration. We
highlight \firstone{best} and \second{second-best} values for each metric. $\heartsuit$ and $\spadesuit$, and denote the all-in-one image restoration and all-in-one video restoration methods, respectively.
}
\vspace{1mm}
\label{tab:all-in-one}
\centering
\resizebox{\textwidth}{!}{%
\setlength{\tabcolsep}{3pt} %
\begin{tabular}{@{}lcccccccccccc}
\toprule
             & \multicolumn{6}{c}{\mytext{Double Degradation}}&\multicolumn{6}{c}{\mytext{Triple Degradation}}\\ \cmidrule(l){2-7} \cmidrule(l){8-13} 
             & PSNR \upcolor{$\uparrow$}  & SSIM \upcolor{$\uparrow$}   & LPIPS \upcolor{$\downarrow$}  & MANIQA\upcolor{$\uparrow$}& CLIP-IQA\upcolor{$\uparrow$}& MUSIQ\upcolor{$\uparrow$}& PSNR \upcolor{$\uparrow$}  & SSIM \upcolor{$\uparrow$}   & LPIPS \upcolor{$\downarrow$}  & MANIQA\upcolor{$\uparrow$}& CLIP-IQA\upcolor{$\uparrow$}& MUSIQ\upcolor{$\uparrow$}\\
$\heartsuit$AirNet~\cite{AirNet}& 15.52& 0.4709& 0.6137& 0.1815& 0.2921& 29.56& 12.70& 0.3408& 0.7958& 0.1420& 0.2941& 22.82\\
$\heartsuit$PromptIR~\cite{potlapalli2023promptir}& 15.84& 0.4553& 0.6363& 0.1903& 0.2956& 29.85& 13.04& 0.3189& 0.8339& 0.1474& 0.2939& 23.01\\
$\heartsuit$DA-CLIP~\cite{daclip}& 16.79& 0.4909& 0.6177& \second{0.2142}& \second{0.3093}& \second{35.27}& 15.66& 0.5254& 0.5603& 0.2196& 0.2632& \second{25.83}\\
$\heartsuit$HAIR~\cite{cao2024hair}& 15.86& 0.4593& 0.6308& 0.1875& 0.2811& 29.76& 13.07& 0.3247& 0.8247& 0.1456& 0.2727& 23.29\\
\cdashline{2-13}
$\spadesuit$BasicVSR++~\cite{chan2022generalization}
& \second{20.45}& 0.6049& 0.5146& 0.2108& 0.2767& 25.65& 17.09& 0.5026& 0.5579& \second{0.2474}& \second{0.3322}& 24.65\\
$\spadesuit$AverNet~\cite{NEURIPS2024_e635a25e}& 19.03& \second{0.6583}& \second{0.3949}& 0.1894& 0.2394& 26.02& \second{20.62}& \second{0.6481}& \second{0.3564}& 0.1556& 0.1907& 15.40\\
\mytext{MoA-VR-Ours}& \firstone{23.47}& \firstone{0.6852}& \firstone{0.3386}& \firstone{0.2238}& \firstone{0.3367}& \firstone{36.14}& \firstone{22.94}& \firstone{0.6497}& \firstone{0.3074}& \firstone{0.2585}& \firstone{0.3156}& \firstone{30.43}\\
\bottomrule
\end{tabular}%
}
\end{table*}

\vspace{1pt}

\begin{algorithm}[t]
\KwIn{Low-quality video $V$}
\KwOut{Restored high-quality video $\tilde{V}$}

\BlankLine
\caption{Agent Collaboration in MoA-VR} 
\label{alg:unified_restoration_workflow}

\LineComment{Identify the type of degradation in the input video}
$degradation \gets \mathcal{A}_i(V)$\;

\LineComment{Route the degradation removal sequence}
$plan \gets \mathcal{A}_r(degradation)$\;

\While{$plan$ is not empty}{
    $subtask \gets \text{First}(plan)$\;
    
    $tools \gets \text{Toolbox}(subtask)$\;
    
    $candidates \gets \emptyset$\;
    
    \ForEach{$tool \in tools$}{
        $V' \gets tool(V)$\;
        
        \LineComment{Evaluate the quality of the restored video to determine a best tool}
        $score \gets \mathcal{A}_a(V', subtask)$\;
        
        Add $(V', score)$ to $candidates$\;
    }

    $(\tilde{V}, best\_score) \gets \text{SelectBest}(candidates)$\;

    $success \gets \mathcal{A}_i(\tilde{V}, subtask)$\;

    \If{success}{
        $V \gets \tilde{V}$\;
        Remove $subtask$ from $plan$\;
    } \Else {
        \LineComment{Reroute if previous restoration fails} \;
        $plan \gets \mathcal{A}_r(plan, subtask)$\;
    }
}

\textbf{output} $V$
\end{algorithm}

Next, the Routing and Restoration Agent $\mathcal{A}_r$ consults a knowledge base and leverages prior restoration experiences to formulate a customized restoration path. This path consists of an ordered sequence of subtasks drawn from the restoration toolbox. For each current subtask, $\mathcal{A}_r$ activates all relevant tools, each generating a candidate restored output. These candidates are then evaluated by the Quality Assessment Agent $\mathcal{A}_a$, which employs a vision-language quality regression model to assess perceptual quality and select the best-performing candidate. The selected output becomes the final result of the current iteration and the input for the next iteration.

The closed-loop nature of MoA-VR is embodied in its iterative feedback mechanism. In each subsequent iteration, $\mathcal{A}_i$ re-assesses the degradation state. If a particular degradation type is no longer detected (i.e., judged as \textit{none}), it signifies the success of the preceding restoration attempt. The Routing and Restoration Agent $\mathcal{A}_r$ then advances along the planned route to apply the next subtask. Conversely, if degradation persists, the system interprets this as a failed restoration step. Consequently, $\mathcal{A}_r$ triggers a rollback and dynamically replans an alternative restoration sequence, as illustrated in Fig.~\ref{fig:route}, selecting a new toolbox for the subtask. The cycle then continues with updated context.

Throughout this iterative process, the three agents synergistically contribute their strengths: $\mathcal{A}_i$ provides precise degradation perception, $\mathcal{A}_r$ enables adaptive decision-making and planning, and $\mathcal{A}_a$ delivers human-aligned quality evaluation. Together, they allow MoA-VR to flexibly and effectively address diverse and complex mixed degradation scenarios, yielding a scalable and generalizable solution for real-world video restoration challenges.

\section{Experiment}
\label{exp}


\subsection{Configurations}
\textbf{Dataset.} We employ two custom-curated datasets tailored to the multi-agent setting: {1) MoA-VD: A multi-degradation video dataset used to train and evaluate the Degradation Identification Agent (\(\mathcal{A}_i\)) and the Routing and Restoration Agent (\(\mathcal{A}_r\)). We construct the dataset with diverse combinations of eight typical distortions (e.g., blur, noise, compression). The training set contains a wide range of degradation permutations, while the test set comprises 400 video clips with degradation combinations not seen during training, ensuring generalization rather than memorization. 2) Res-VQ: A quality-labeled video dataset built for evaluating the Quality Assessment Agent (\(\mathcal{A}_a\)). It includes both degraded and restored video sequences with MOS as ground truth. An 80:20 split is applied for training and evaluation.


\textbf{Implementation.}  
All experiments are conducted on NVIDIA GeForce RTX 3090 and RTX A40 GPUs. The degradation identification agent (\(\mathcal{A}_i\)) is built upon Qwen2.5-VL-7B~\cite{Qwen2.5-VL}, fine-tuned for 5 epochs using the AdamW optimizer with a learning rate of \(1 \times 10^{-3}\), a batch size of 4, a LoRA rank of 16, and a LoRA alpha of 32. The routing and restoration agent (\(\mathcal{A}_r\)) integrates a set of task-specific state-of-the-art models: VRT~\cite{liang2022vrt}, AverNet~\cite{NEURIPS2024_e635a25e}, and DSTNet~\cite{Pan_2023_CVPR} for video denoising, decompression, and deblurring; FastLLVE~\cite{li2023fastllve} and RQLLIE~\cite{Liu_2023_ICCV} for low-light video enhancement; MAP-Net~\cite{xu2023map} and Dehazeformer~\cite{song2023vision} for video dehazing; Rainmamba~\cite{wu2024rainmamba} for video deraining; GIMMVFI~\cite{guo2024generalizable} for low-frame-rate video interpolation; and BasicVSR++~\cite{chan2022generalization} and Real-ESRGAN~\cite{wang2021realesrgan} for video super-resolution. For the quality assessment agent (\(\mathcal{A}_a\)), we adopt InternLM-8B~\cite{team2023internlm} as the base model, which is trained with AdamW for 10 epochs using a batch size of 8 and a LoRA rank of 16.

\subsection{Comparison}
To validate the effectiveness of our proposed MoA-VR framework, a flexible and scalable mixture-of-agent system for all-in-one video restoration, we compare it against six representative state-of-the-art methods, including four image-based all-in-one restoration approaches and two video restoration methods designed for mixed distortion scenarios. Notably, while image-based methods are typically trained on a wider variety of degradation types, video-based approaches such as BasicVSR++ are generally limited to handling only two distortion types simultaneously. This limitation reveals a crucial gap in existing video restoration techniques when faced with complex, multi-type distortions, which MoA-VR aims to address comprehensively. All comparative methods are evaluated using the official pre-trained models and code provided by their authors to ensure fairness.

\begin{figure*}[tp]
    \centering
    \includegraphics[width=1\linewidth]{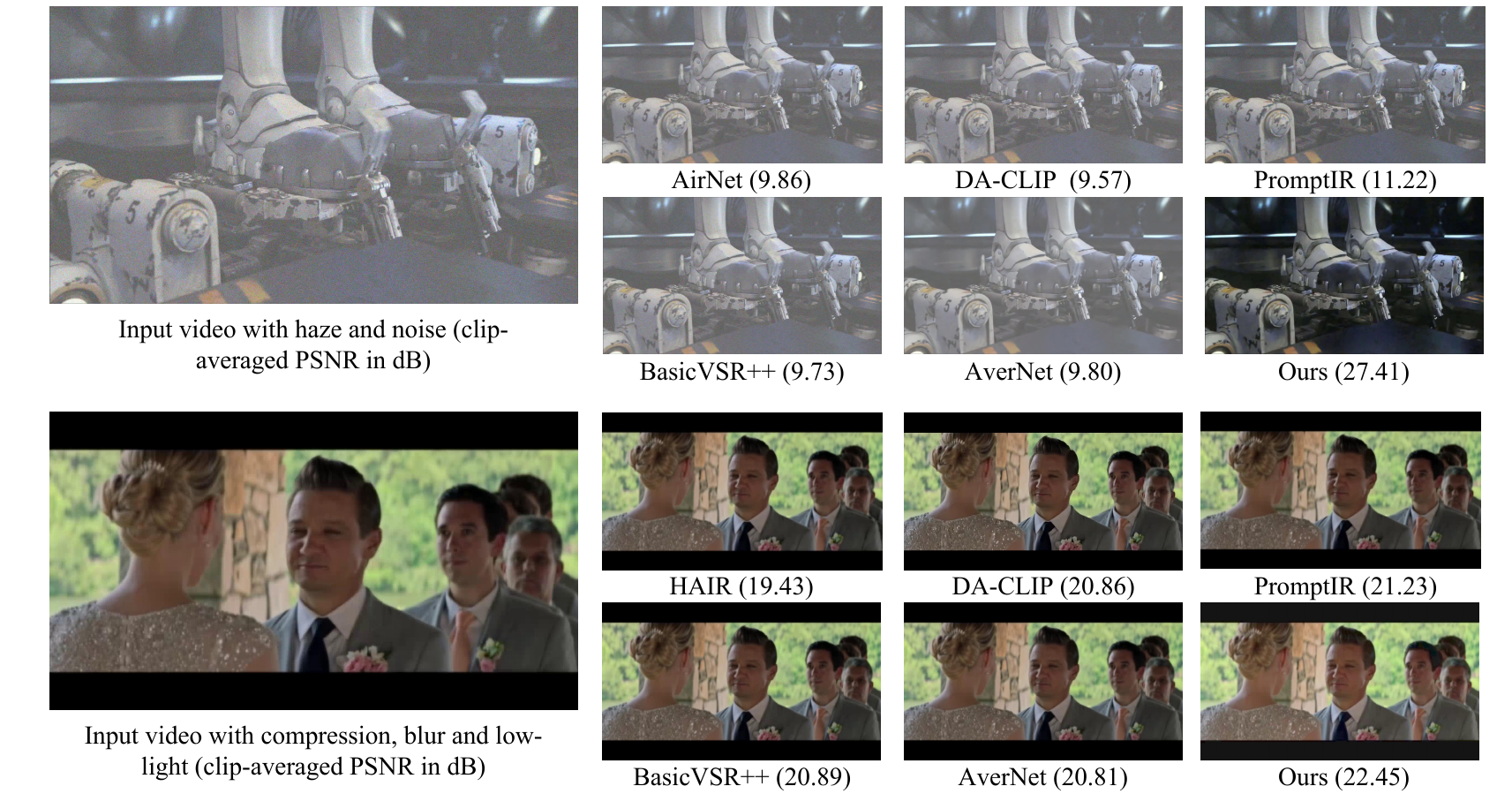}
    \caption{Performance comparison of MoA-VR with other all-in-one methods, using the first frame as an illustration. }
    \label{fig:visual}
\end{figure*}

\textbf{Quantitative comparisons.} Table~\ref{tab:all-in-one} presents quantitative results where MoA-VR consistently outperforms all existing all-in-one image and video restoration methods across both pixel-level (PSNR, SSIM) and perceptual quality metrics (LPIPS~\cite{lpips}, MANIQA~\cite{maniqa}, CLIP-IQA~\cite{clipiqa}, MUSIQ~\cite{musiq}). Under double degradation, MoA-VR achieves a significant PSNR of 23.47 dB, outperforming the strongest baseline BasicVSR++ by \textbf{3.02} dB, and improves LPIPS by 14.2\%. When restoring videos containing three types of distortions, the advantages and robustness of our method become even more evident. Compared to BasicVSR++, the PSNR gain increases to \textbf{5.85} dB, and it still outperforms all other methods.


\textbf{Qualitative comparisons.} These solid quantitative results are visually corroborated by our qualitative comparisons (Fig.~\ref{fig:visual}), which reveal MoA-VR's exceptional ability to preserve fine details and effectively suppress artifacts under demanding mixed distortion conditions. Unlike existing all-in-one image restoration techniques such as DA-CLIP, which may falter in video restoration due to differing degradation characteristics, or contemporary all-in-one video restoration models like BasicVSR++, which struggle with degradations such as haze and low-light not addressed in their design, MoA-VR effectively restores a wide array of degradations, yielding visually superior results. 



These results demonstrate three main advantages of MoA-VR: (1) The collaborative operation of agents \(\mathcal{A}_i\), \(\mathcal{A}_r\), and \(\mathcal{A}_a\) enables effective handling of complex, multi-type degradations, delivering state-of-the-art restoration performance. (2) MoA-VR exhibits remarkable scalability, supported by an extensible restoration toolbox, an adaptive routing strategy, and robust distortion identification coupled with quality assessment. It is the first all-in-one video restoration method capable of addressing eight types of mixed distortions, surpassing existing methods in versatility. (3) MoA-VR sets a new paradigm for multi-agent collaboration in video restoration, facilitating easy extension and optimization to diverse real-world scenarios by leveraging modular agent cooperation.
\begin{table*}[t]
\caption{
Comparison of MoA-VR with all-in-one methods for complex video restoration.
We highlight the \firstone{best} and \second{second-best} values for each metric.
}
\vspace{-1mm}
\label{table3}
\centering
\renewcommand{\arraystretch}{1.2} 
\resizebox{0.7\textwidth}{!}{%
\setlength{\tabcolsep}{3pt} %
\begin{tabular}{@{}lcccccccc}
\toprule
& \multicolumn{4}{c}{\mytext{Multi-Order Degradation}}& \multicolumn{4}{c}{\mytext{Real-World Degradation}}\\
\cmidrule(l){2-5} \cmidrule(l){6-9}
& MANIQA\upcolor{$\uparrow$} & CLIP-IQA\upcolor{$\uparrow$} & MUSIQ\upcolor{$\uparrow$} & TOPIQ\upcolor{$\uparrow$} & MANIQA\upcolor{$\uparrow$} & CLIP-IQA\upcolor{$\uparrow$} & MUSIQ\upcolor{$\uparrow$} & TOPIQ\upcolor{$\uparrow$} \\

BasicVSR++~\cite{chan2022generalization} 
& \second{0.1708} & \second{0.2713} & \second{27.80} & \second{0.3233} & \second{0.2054} & 0.2564 & \second{42.35} & \second{0.3533} \\
AverNet~\cite{NEURIPS2024_e635a25e}     
& 0.1569 & 0.2485 & 23.45 & 0.3091 & 0.2001 & \second{0.3028} & 41.15 & 0.3451 \\
\mytext{MoA-VR (Ours)}                  
& \firstone{0.1801} & \firstone{0.2816} & \firstone{38.41} & \firstone{0.3556} & \firstone{0.2238} & \firstone{0.3108} & \firstone{43.31} & \firstone{0.3611} \\
\bottomrule
\end{tabular}%
}
\end{table*}

\begin{figure*}[ht]
    \centering
\includegraphics[width=0.9\linewidth]{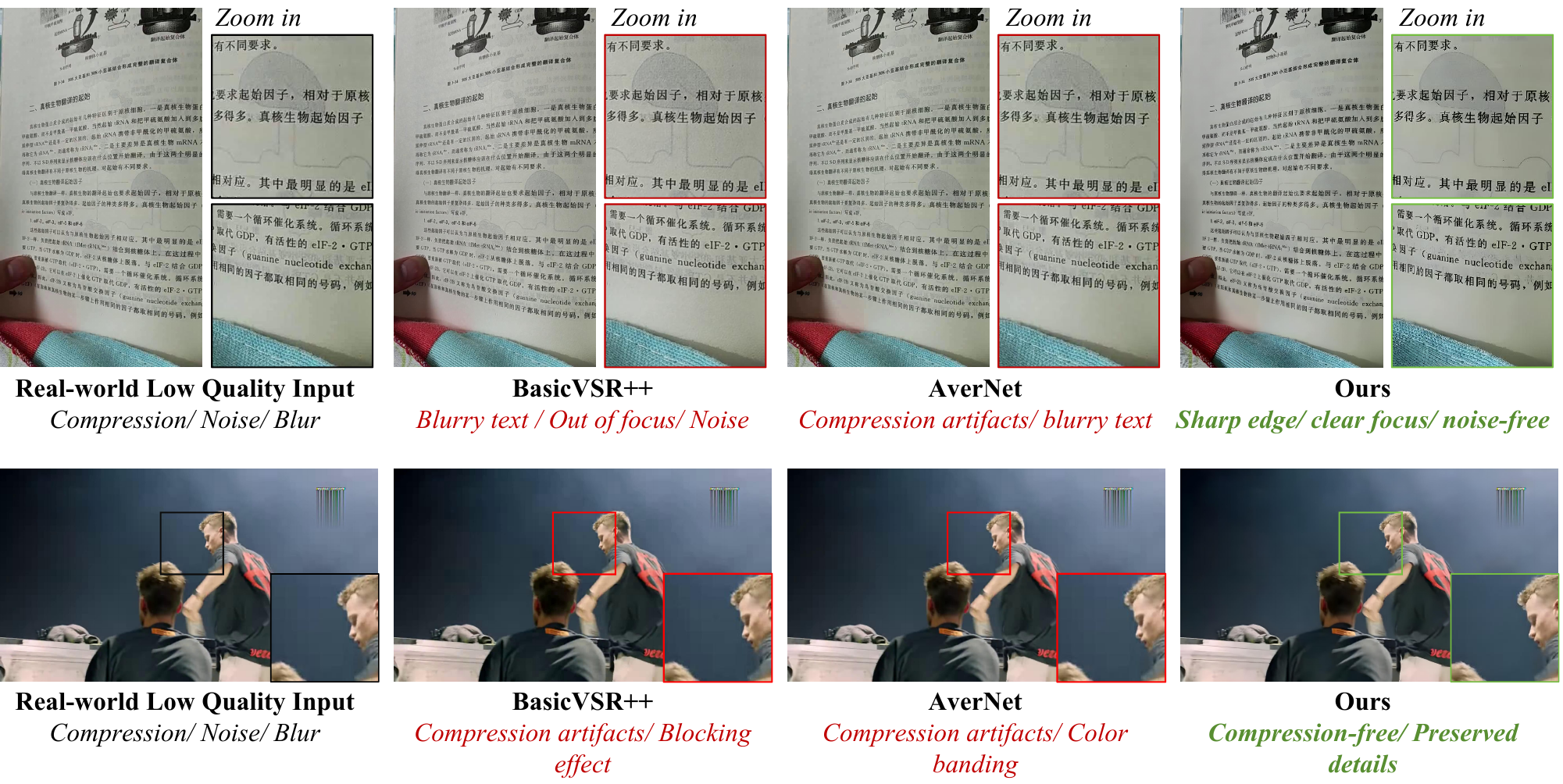}
    \caption{\textcolor{black}{Quantitative comparison of MoA-VR with other all-in-one video restoration methods on real world video restoration, using first frame as an illustration.}}
    \label{fig:visualrw}
\end{figure*}
\textcolor{black}{\textbf{Comparison on Complex Degradations.} We conducted additional experiments to validate the generalization of our model for real-world videos or multi-order degraded videos. Specifically, we generated a synthetic multi-order test set containing 50 videos, and additionally collected 50 real-world videos from existing video restoration datasets~\cite{duan2024finevq}. }

\textcolor{black}{Table~\ref{table3} presents quantitative results where MoA-VR consistently outperforms all existing all-in-one video restoration methods across four perceptual quality metrics (MANIQA~\cite{maniqa}, CLIP-IQA~\cite{clipiqa}, MUSIQ~\cite{musiq}, TOPIQ~\cite{chen2024topiq}). Under multi-order degradation, MoA-VR improves MUSIQ~\cite{maniqa} by \textbf{40\%}. When restoring real-world videos, the advantages of our method is evident as well, with improvement across all quality metrics compared with BasicVSR++~\cite{chan2022generalization}.}

\textcolor{black}{Fig.~\ref{fig:visualrw} presents quantitative comparison, which reveal MoA-VR's robustness on real-world videos. Compared with other all-in-one restorers, MoA-VR handles real-world scenarios more robustly due to its integration of various tools. It removes compression artifacts, noise, and blur in a step-by-step manner, while others typically solve single degradation, leaving residual compression artifacts (e.g. AverNet~\cite{NEURIPS2024_e635a25e}) or noise (e.g. BasicVSR++~\cite{chan2022generalization}). }

\textcolor{black}{Both quantitative and quantitative results demonstrate that our method generalizes robustly to more complex multi-order and real-world degradations.}

\subsection{Evaluation}
\textbf{Degradation Identification Accuracy.} To assess both the zero-shot capability of VLMs in degradation recognition, we benchmark several open-source VLMs (e.g., Qwen, and LLaVA) on the classification task using MoA-VD. We adopt accuracy per degradation type and overall average accuracy as the evaluation metric. As shown in Table~\ref{tab:cls}, VLMs exhibit strong generalization in identifying complex, mixed degradations from multimodal inputs.  Among all evaluated models, the Qwen series consistently outperforms others, achieving an average accuracy of \textbf{51.18\%} on degradation level prediction. Given its superior performance, we fine-tune Qwen on the MoA-VD training set and adopt it as the backbone for our degradation identification agent \(\mathcal{A}_i\). After fine-tuning, our (\(\mathcal{A}_i\)) achieves an average accuracy of \textbf{87.2\%} across all degradation types. Specifically, it performs best  on blur (95.5\%), haze (95.0\%), and noise (94.0\%), while maintaining expressive accuracy on artifact-heavy degradations like low-resolution and low-light ($\geq$80\%). 
\begin{table*}[tp]
\vspace{-0.3em}
\centering
\caption{Benchmark of state-of-the-art VLMs and the proposed degradation identification agent on degradation level prediction task. $\diamondsuit$, and $\heartsuit$ denote the zero-shot VLMs and fine-tuned VLMs, respectively. The best results are highlighted in \firstone{best}, and the second-best results are highlighted in \second{second-best}. The evaluation metric is prediction accuracy.}
\begin{tabular}{l|cccccccc|c}
\toprule
\bf \mytext{Model} & \bf \mytext{Noise} & \bf \mytext{Compre.} & \bf \mytext{Blur} & \bf \mytext{Low Light} & \bf \mytext{Rain  } & \bf \mytext{Haze} & \bf \mytext{Low Res.} & \bf \mytext{Low Fra.} & \bf \mytext{Avg.}\\
\midrule
InternVL2 (1B) \cite{chen2024expanding}
& 0.3800 
& 0.5150 
& 0.5575 
& 0.2600 
& 0.5325
& 0.4300 
& 0.3550 
& 0.4125
& 0.4303
\\
InternVL2 (2B) \cite{chen2024expanding}
& 0.5250
& 0.4925
& 0.4775
& 0.4050
& 0.5825
& 0.4350
& 0.4025
& 0.5850
& 0.4881
\\
InternVL3 (1B) \cite{wang2024mpo}
& 0.5375 
& 0.5100 
& 0.4050 
& \second{0.4900}& 0.5875 
& 0.3375 
& 0.4125 
& 0.4400
& 0.4650
\\
LLaVA-NEXT-Video (7B) \cite{liu2024llavanext}
& 0.4075 
& 0.2550 
& 0.3800 
& 0.2600 
& 0.3350 
& 0.2750 
& 0.3450 
& 0.2100
& 0.3084
\\
mPLUG-Owl3 (1B) \cite{ye2024mplug}
& 0.3750 
& 0.2500 
& 0.4475 
& 0.2700 
& 0.5200 
& 0.2500 
& 0.3150 
& 0.1875
& 0.3269
\\
mPLUG-Owl3 (7B) \cite{ye2024mplug}
& \second{0.6450}& 0.4600 
& 0.5175 
& 0.4400 
& 0.5550 
& 0.4925 
& 0.4350 
& 0.5025
& 0.5059
\\
Qwen2.5VL (3B) \cite{Qwen2.5-VL}
& 0.5250 
& 0.2625 
& \second{0.5900}& 0.2900 
& 0.5525 
& 0.4175 
& \second{0.6450} 
& \second{0.6825}
& 0.4956
\\ 
Qwen2.5VL (7B) \cite{Qwen2.5-VL}
& 0.5700 
& 0.4150 
& 0.5450 
& 0.3950 
& \second{0.6500} 
& \second{0.4650} 
& 0.5750 
& 0.4800
& \second{0.5118}
\\


\hdashline
\mytext{\textbf{$\mathcal{A}_{i}$}-Ours}
& \firstone{0.9400} 
& \firstone{0.8600} 
& \firstone{0.9550} 
& \firstone{0.7525} 
& \firstone{0.8850} 
& \firstone{0.9500} 
& \firstone{0.8000} 
& \firstone{0.8350}
& \firstone{0.8722} \\
\bottomrule
\end{tabular}

\label{tab:cls}
\vspace{-0.4cm}
\end{table*}

\begin{table*}[t]
\caption{
Comparison of our routing strategy with other degradation removal strategies for multi-degraded video restoration. We
highlight \firstone{best} and \second{second-best} values for each metric. 
}
\vspace{1mm}
\label{tab:route}
\centering
\resizebox{\textwidth}{!}{%
\setlength{\tabcolsep}{3.5pt} %
\begin{tabular}{@{}lcccccccccccc@{}}
\toprule
& \multicolumn{6}{c}{\mytext{Double Degradation}}         & \multicolumn{6}{c}{\mytext{Triple Degradation}}     \\ \cmidrule(l){2-7} \cmidrule(l){8-13} 
& PSNR \upcolor{$\uparrow$}  & SSIM \upcolor{$\uparrow$}   & LPIPS \upcolor{$\downarrow$}   & MANIQA\upcolor{$\uparrow$}& CLIP-IQA\upcolor{$\uparrow$}& MUSIQ\upcolor{$\uparrow$}& PSNR \upcolor{$\uparrow$}  & SSIM \upcolor{$\uparrow$}   & LPIPS \upcolor{$\downarrow$}  & MANIQA\upcolor{$\uparrow$}& CLIP-IQA\upcolor{$\uparrow$}& MUSIQ\upcolor{$\uparrow$}\\
Reverse Order& 18.24& 0.6181& 0.4719& 0.1813& 0.2405& 27.72& 17.32& 0.5914& 0.5102& 0.1347& 0.2124& 22.56\\
Random Order&19.63& 0.6208& 0.4139& 0.1857& 0.2566& 25.64& 18.94& 0.6150& 0.4523& 0.1482& 0.2203& 23.87\\
Expert Order& 19.89& 0.6340& 0.3915& 0.1855& 0.3047& 33.73& 19.75& 0.6321& 0.4010& 0.1860& 0.2787& 27.25\\
$\mathcal{A}_r$-Zero-Shot& 20.31& 0.6382& 0.3798& 0.1914& 0.3125& 34.02& 20.12&0.6367& 0.3894& 0.1913& 0.2776&27.92\\
$\mathcal{A}_r$-Experience& \second{20.98}& \second{0.6415}& \second{0.3687}& \second{0.1976}& \second{0.3198}& \second{34.89}& \second{20.23}& \second{0.6442}& \second{0.3721}& \second{0.1928}& \second{0.2842}& \second{28.11}\\
\mytext{$\mathcal{A}_r$-Ours}& \firstone{21.65}& \firstone{0.6463}& \firstone{0.3544}& \firstone{0.2042}& \firstone{0.3263}& \firstone{35.43}& \firstone{21.11} & \firstone{0.6483}& \firstone{0.3713}& \firstone{0.2081} & \firstone{0.3012}&\firstone{29.74}\\
\bottomrule
\end{tabular}%

}
\end{table*}
\textbf{Routing and Restoration Strategy.} To examine how the order of degradation removal affects final restoration quality, we design and evaluate six routing strategies. These include reverse-order restoration, random restoration, and a fixed expert-defined order, as well as three variants of our Routing and Restoration Agent (\(\mathcal{A}_r\)). The first three serve as baselines, ranging from naive to human-guided sequences, while the latter three evaluate the contributions of learned experience and rollback. 

Specifically, \(\mathcal{A}_r\)-Zero-Shot leverages LLM predictions without any prior experience or rollback; \(\mathcal{A}_r\)-Experience Only builds on training-set feedback to predict restoration sequences but cannot dynamically correct mistakes; and \(\mathcal{A}_r\)-Ours integrates both experience learning and rollback to actively refine routing decisions. The expert-defined order consists of a set of static sequences, each tailored for a specific combination of degradation types. Table~\ref{tab:route} shows the quantitative results. Note that all other model parameters and settings are held constant across experiments to ensure fair comparison.

The results reveal several important insights. Reverse-order restoration performs worst, confirming that the degradation process is not trivially reversible. Expert-defined ordering improves over naive strategies but remains suboptimal under mixed degradations due to its rigidness. Notably, the zero-shot LLM-based routing already surpasses handcrafted rules, demonstrating the potential of data-driven sequence planning.  Furthermore, incorporating experience from training data leads to further gains.

\textcolor{black}{Random and Reverse restoration orders yield the worst results, indicating that restoration order has a significant impact on final quality, and an incorrect order can leave degradations unremoved or even amplify them. Although the Expert Order outperforms random and reverse orders, it is still notably inferior to the automatic routing strategies, showing that fixed human-crafted orders are less effective than data-driven dynamic adjustments. Adapting the order to each video’s specific degradation profile consistently outperforms any fixed order. Among the three dynamic-order variants, \(\mathcal{A}_r\)-Ours consistently surpasses \(\mathcal{A}_r\)-Experience, demonstrating that an “initial planning + real-time adjustment” approach is more capable of handling diverse degradation combinations and generalizes better than relying solely on past experience.
In the Triple Degradation setting, the improvement in perceptual quality metrics is even larger than in the Double Degradation case, highlighting that the routing strategy becomes increasingly critical as degradation complexity increases.}

Finally, when introducing both experience based routing and roll-back based rerouting, our full agent \(\mathcal{A}_r\)-Ours achieves the highest performance across all metrics. For instance, compared with non-intelligent methods, it yields a PSNR improvement of \textbf{19.89+1.76} dB and SSIM gain of \textbf{0.634+0.01} over the expert-defined strategy on the test set. The accumulated experience enables the agent to learn globally optimal routing patterns, while rollback acts as a corrective mechanism that prevents the model from being trapped in suboptimal sequences. This combination proves essential in adapting to diverse and complex degradation scenarios, ultimately leading to superior restoration fidelity.

\begin{table}
\vspace{-0.3em}
\centering
\caption{\textcolor{black}{Performance of state-of-the-art models and the proposed quality assessment agent on our established Res-VQ database in terms of quality prediction. $\spadesuit$, $\diamondsuit$, and $\heartsuit$ denote the traditional IQA methods, traditional VQA methods, and DNN-based VQA metrics, respectively. It should be noted that all the DNN-based models are re-trained with their default settings. The best results are highlighted in \firstone{best}, and the second-best results are highlighted in \second{second-best} .}}
\vspace{-0.8em}
   \begin{tabular}{lcccc}
    \toprule[1pt]
 \bf \mytext{Methods / Metrics}
&  \bf \mytext{SRCC\upcolor{$\uparrow$}}&  \bf \mytext{KRCC}\upcolor{$\uparrow$}&  \bf \mytext{PLCC}\upcolor{$\uparrow$}&   \bf \mytext{RMSE}\upcolor{$\downarrow$}\\
    \midrule
    $\spadesuit$ NIQE\cite{niqe}&0.3196 &0.2222 &0.1475& N/A  \\
    $\spadesuit$ QAC\cite{qac}&0.0501 &0.0317 &0.0046 & 20.18\\
    $\spadesuit$ HOSA\cite{hosa}&0.3699&0.2584&0.3259 &23.36\\ \hdashline
    $\diamondsuit$ BMPRI\cite{quality:BMPRI}&0.2872&0.2056&0.3339&28.32\\
    $\diamondsuit$ VIDEVAL\cite{tu2021ugc}
    &0.7137&0.5387&0.7586&9.66\\
    $\diamondsuit$ RAPIQUE\cite{tu2021rapique}
    &0.8166&0.6284&0.8402& 8.41\\ \hdashline
    $\heartsuit$ DOVER*~\cite{wu2023dover}&0.7751&0.5787&0.7589&19.20  \\
    $\heartsuit$ SimpleVQA*~\cite{sun2022deep}&0.8007&0.6032&0.7844&13.81 \\
    $\heartsuit$ VSFA\cite{li2019quality}&0.7576  &0.5612&0.7196&11.03 \\
    $\heartsuit$ FAST-VQA~\cite{wu2022fast}&0.7592&0.5554&0.7160&11.09\\
    $\heartsuit$ DOVER ~\cite{wu2023exploring}&0.7877&0.5834&0.7669&10.06\\
    $\heartsuit$ SimpleVQA~\cite{sun2022deep}&0.8596  &0.6671&0.8593&7.85  \\
    $\heartsuit$ GSTVQA~\cite{chen2021learning}&\second{0.8902} &\second{0.7144}&\second{0.8938}&\second{7.62}  \\
   \hdashline
   \mytext{\textbf{\(\mathcal{A}_a\)}-Ours} &\firstone{0.9165}&\firstone{0.7510}&\firstone{0.9284}&\firstone{5.65}\\
    \bottomrule[1pt]
  \end{tabular}
  \label{tab:qa}
\end{table}

\textbf{Quality Assessment Performance.} To evaluate the effectiveness of our proposed quality assessment agent $\mathcal{A}_a$, we conduct comprehensive experiments on the Res-VQ dataset. We benchmark $\mathcal{A}_a$ against 11 representative no-reference VQA models, categorized into: (1) traditional NR-IQA methods (NIQE~\cite{niqe}, QAC~\cite{qac}, HOSA~\cite{hosa}), (2) traditional NR-VQA models (TLVQA~\cite{korhonen2019two}, VIDEVAL~\cite{tu2021ugc}, RAPIQUE~\cite{tu2021rapique}), and (3) deep learning-based NR-VQA methods (VSFA~\cite{li2019quality}, GSTVQA~\cite{chen2021learning}, SimpleVQA~\cite{sun2022deep}, Fast-VQA~\cite{wu2022fast}, Dover~\cite{wu2023exploring}). All models are retrained on Res-VQ with an 80:20 training/testing split. Evaluation metrics include SRCC, PLCC, KRCC, and RMSE, assessing correlation with human MOS scores and prediction accuracy.

\textcolor{black}{As shown in Table~\ref{tab:qa}, traditional NR-IQA models exhibit poor alignment with human perception on restored video quality (e.g., NIQE: SRCC = 0.320), mainly due to the lack of temporal modeling. NR-VQA models offer moderate improvements (e.g., VIDEVAL: SRCC = 0.713), but they still struggle to handle diverse restoration artifacts. Deep learning-based methods such as Simple-VQA and GSTVQA also perform poorly in a zero-shot setting (SRCC = 0.80 and 0.77, respectively), largely because these models are predominantly trained on User-Generated Content (UGC) videos, which differ substantially from the intermediate restored outputs of our agent-based system. Consequently, their generalization to restored video quality assessment is limited. After being trained on our constructed Res-VQ dataset, deep learning-based models demonstrate significant improvements (Simple-VQA: SRCC = 0.859; GSTVQA: SRCC = 0.890), which validates the importance of building a dedicated database for restored video quality assessment.}

Building on this, our agent $\mathcal{A}_a$ achieves the highest performance across all metrics (SRCC = \textbf{0.9165}, PLCC = \textbf{0.9284}, KRCC = \textbf{0.7510}, RMSE = \textbf{5.65}). The results highlight the superior capability of LLM-enhanced architectures in fusing visual and textual cues to assess quality from a human-centric perspective. By leveraging the semantic richness of prompt-aware inputs and the cross-modal reasoning ability of the LLM backbone, $\mathcal{A}_a$ effectively aligns predicted scores with human opinions. Moreover, it enables quality-aware routing guidance in our restoration pipeline, providing not only accurate scoring but also practical value in downstream decision-making.

\textcolor{black}{\textbf{Ablation Study on Degradation Identification.} To assess the generalization ability of the identification agent, we performed additional experiments by modifying the degradation generation pipeline. We constructed an additional test set by replacing all degradation generation procedures so that they differ from those used in the original MoA-VD test set (Group A). Specifically, we introduced impulse noise, motion blur, H.265 compression, and adopted the pipeline of~\cite{jarvisir2025} to synthesize low-light and rain. For low resolution and low frame rate, we randomly varied their scaling factors. Degradations were applied in both single-order and multi-order combinations, yielding 400 new test samples (denoted as Group B). As seen in Fig.~\ref{fig:5-1} and Table \ref{tab:res1-2}, on this unseen dataset, our agent achieved an average recognition accuracy of 84\%. In particular, the recognition accuracy for noise and blur exceeded 90\%, while low-resolution was more easily confused with blur, and low-light performance varied depending on video capture conditions. These results indicate that our degradation identification agent learns generalizable degradation features rather than merely memorizing patterns, and remains robust when facing various degradations.}
\begin{figure}[t]
    \centering
    \includegraphics[width=0.7\linewidth]{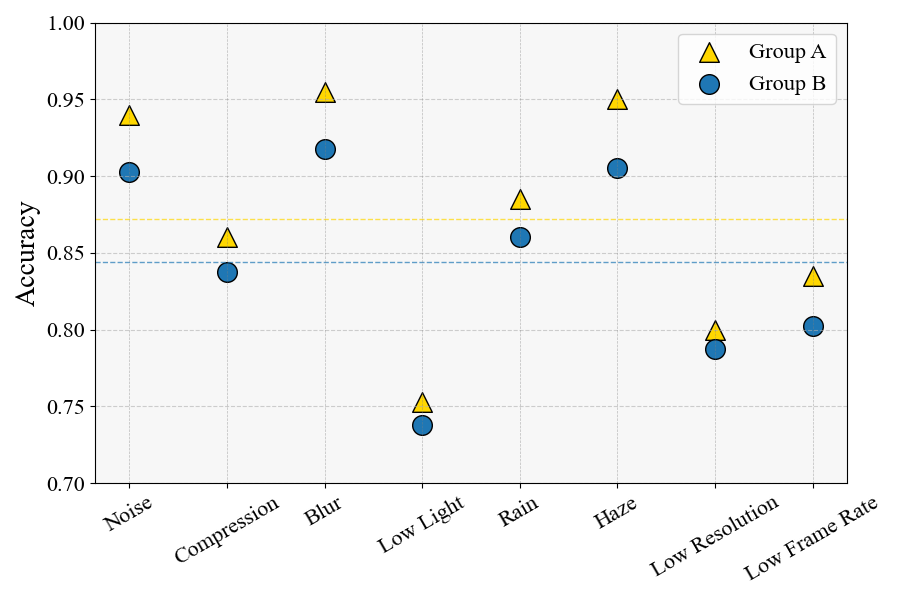}
    \caption{\textcolor{black}{Ablation study on degradation identification. Group A is the original test set and Group B is an additional test set generated with an altered degradation pipeline.}}
    \label{fig:5-1}
\end{figure}
\begin{table}[t]

\centering
\vspace{-10pt}
\caption{Prediction Accuracy of the degradation identification agent across degradation types. Group A uses the original test set (in-domain), while Group B denotes out-of-domain data. higher is better.}
\resizebox{\linewidth}{!}{
\begin{tabular}{l|cccccccc|c}
\toprule
\bf \mytext{Dataset} & \bf \mytext{Noise} & \bf \mytext{Compre.} & \bf \mytext{Blur} & \bf \mytext{Low Light} & \bf \mytext{Rain  } & \bf \mytext{Haze} & \bf \mytext{Low Res.} & \bf \mytext{Low Fra.} & \bf \mytext{Avg.}\\
\midrule
Group A (In-domain)
&0.9400& 0.8600& 0.9550& 0.7525& 0.8850& 0.9500& 0.8000& 0.8350&0.8722 \\
Group B (Out-of-domain)
&0.9025& 0.8375& 0.9175& 0.7375& 0.8600& 0.9050& 0.7875& 0.8025&0.8438\\
\bottomrule
\end{tabular}
}
\label{tab:res1-2}

\end{table}
}

\begin{figure*}
    \centering
    \includegraphics[width=1\linewidth]{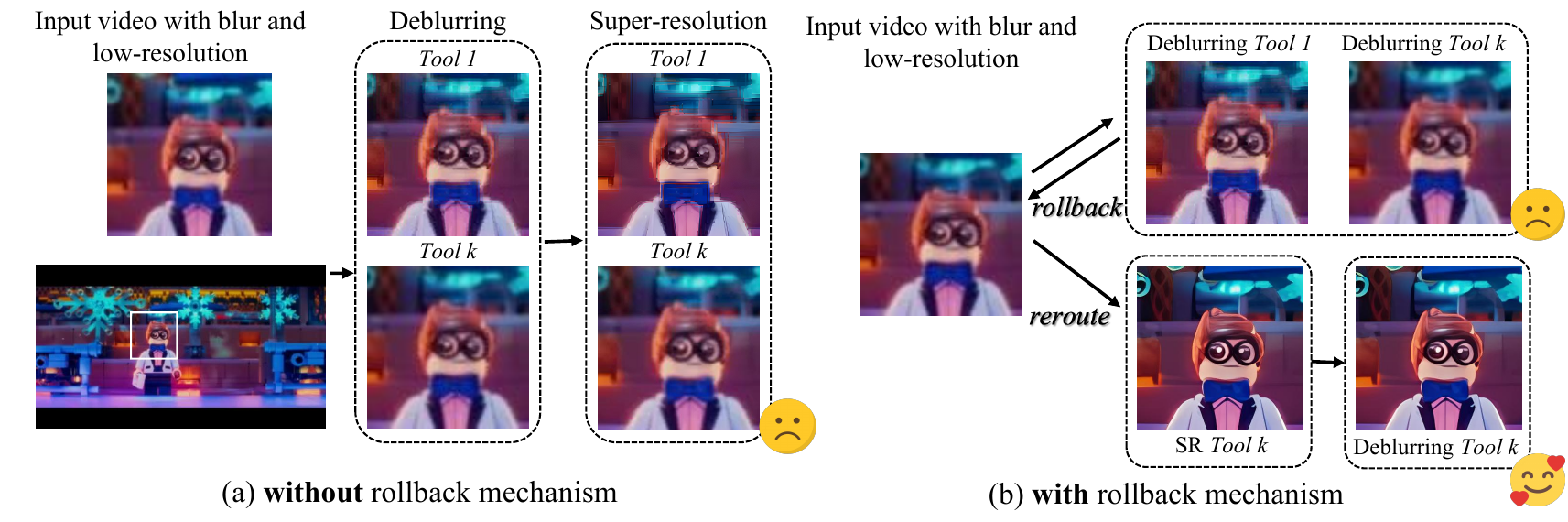}
    \caption{Ablation study on the rollback mechanism. (a) Without rollback, the agent is unable to revise suboptimal decisions, leading to degraded results. (b) Enabled by rollback, the agent re-evaluates and adjusts its restoration route, producing clearer and more accurate outputs.}
    \label{fig:rollback}
\end{figure*}

\begin{figure}
    \centering
    \includegraphics[width=1\linewidth]{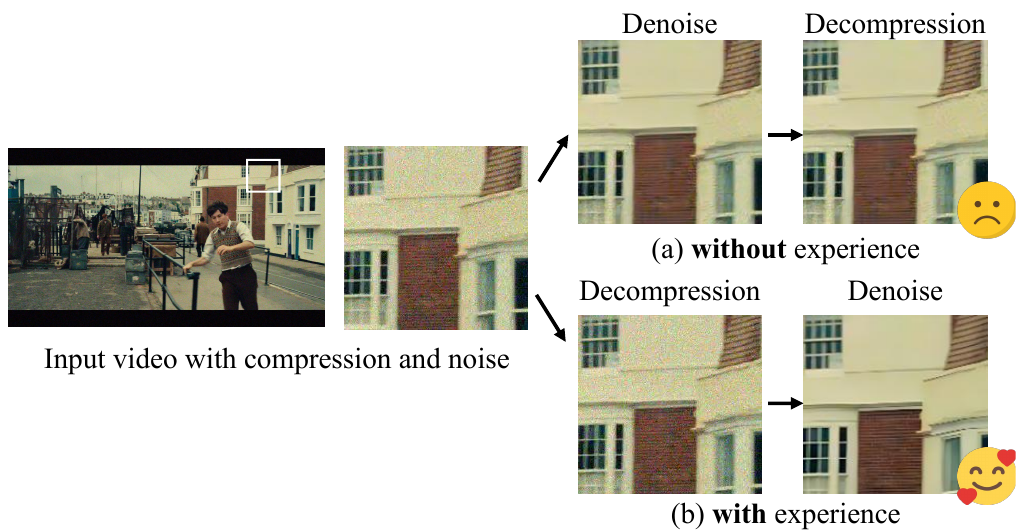}
    \caption{Ablation study on the experience mechanism. (a) Without experience guidance, the agent misorders the operations, resulting in significant quality degradation. (b) Leveraging the experience mechanism, the agent selects an effective restoration sequence, yielding improved perceptual quality and structure preservation.}
    \label{fig:experience}
\end{figure}

\textbf{Ablation Study on Rollback and Experience.} To validate the effectiveness of key components in our MoA-VR framework, we conduct ablation studies on the experience mechanism and the rollback mechanism, as illustrated in Fig.~\ref{fig:experience} and Fig.~\ref{fig:rollback}. 

Without the experience mechanism, the agent selects restoration sequences purely based on its innate policy, which often results in suboptimal decisions. For instance, in Fig.~\ref{fig:experience}(a), the agent mistakenly applies denoising before decompression, treating compression artifacts as noise and causing over-smoothing and amplified residuals. In contrast, Fig.~\ref{fig:experience}(b) demonstrates that with the experience mechanism, the agent first performs decompression followed by denoising, yielding improved perceptual quality and structural fidelity. 

Similarly, the rollback mechanism is crucial for correcting poor routing decisions even when guided by experience. Fig.~\ref{fig:rollback}(a) shows that performing deblurring before super-resolution, despite mild blur, induces over-smoothing, compromising the effectiveness of super-resolution. With the rollback mechanism (Fig.~\ref{fig:rollback}(b)), the agent receives degradation feedback, rolls back, and reroutes to super-resolution first, resulting in better detail preservation. 

These results underscore the necessity of both mechanisms. The experience mechanism facilitates better initial decision-making based on prior observations, while the rollback mechanism provides the flexibility to revise routes dynamically. This dual capability empowers MoA-VR with expert-like adaptability, critical for robust restoration across diverse real-world degradations.

\textcolor{black}{\textbf{Cross-dataset Validation of Quality Assessment Agent.} We conduct cross-dataset validation to assess generalization ability of our quality assessment agent. To the best of our knowledge, Res-VQ is the first VQA dataset specifically designed for restored videos. Hence, we choose existing UGC VQA database as validation. We train SimpleVQA~\cite{sun2022deep}, GSTVQA~\cite{wu2022fast}, and our quality assessment agent on Res-VQ-train and evaluate on Fine-VD-test~\cite{duan2024finevq}, reporting SRCC, PLCC, KRCC, and RMSE. As shown in Table~\ref{table5}, our quality assessment agent consistently outperforms state-of-the-art VQA methods (e.g., \textbf{+0.23} SRCC and a \textbf{13.6} RMSE reduction relative to SimpleVQA~\cite{sun2022deep}). This indicates that our quality assessment agent exhibits strong generalization to novel content not seen during training. }
\begin{table}[t]

\caption{\textcolor{black}{The cross-dataset evaluation results. The models are trained on our Res-VQ and tested on other UGC VQA dataset.}}
\centering
\resizebox{\linewidth}{!}{
\begin{tabular}{lcccc}
   \toprule
 \textit{Test on:} &  \multicolumn{4}{c}{Fine-VD}  \\  \cmidrule(lr){2-5} 
 \textit{Train on:} Res-VQ   & SRCC$\uparrow$&PLCC$\uparrow$ & KRCC$\uparrow$&RMSE$\downarrow$\\  
     \midrule
SimpleVQA~\cite{sun2022deep} &0.3597&0.3305& 0.2493&24.82\\
   GSTVQA~\cite{chen2021learning} &0.3345&0.3238& 0.2251&18.83\\ 
  
\rowcolor{gray!20} \textbf{Ours}   &\textbf{0.5884}&\textbf{0.6211}& \textbf{0.4195}&\textbf{11.22}\\ \bottomrule
  \end{tabular}}
  \label{table5}
  \end{table}

\textbf{Time Complexity Analysis.} We analyze the computational complexity of different routing strategies.
For a video with $n$ degradation types,

 (1) Full Search (Exhaustive): Full search strategy evaluates all possible restoration orders. Hence, the time complexity of full search is
\begin{equation}
T_{\text{full search}} = O(n \times n!)
\end{equation}
which grows factorially with the number of degradations and becomes intractable for large $n$. 

(2) Tree Search: 
Tree search improves upon full search by eliminating the outer $n$ factor. Its complexity can be expanded as
\begin{equation}
\begin{aligned}
T_{\text{tree search}} 
&= O\left(n! \sum_{k=0}^{n-1} \frac{1}{k!}\right) \\[6pt]
\end{aligned}
\end{equation}
As $n \to \infty$, the summation 
\[
\sum_{k=0}^{n-1} \frac{1}{k!} \;\longrightarrow\; e,
\]
which is a constant. Therefore, the asymptotic time complexity is
\begin{equation}
T_{\text{tree}} = O(n!).
\end{equation}

(3) Ours:
Building on tree search, our method leverages an LLM-based predictor to propose a highly probable restoration sequence.
At each step, the predictor succeeds with probability $p \in (0,1]$.
When a wrong decision is made, rollback occurs, leading to an expected cost of $(1-p)\cdot i$ at depth $i$.
Summing over all $n$ steps yields an overall expected complexity of
\begin{equation}
T_{\text{ours}} = O!\left(\frac{n}{p} + (1-p)n^2\right),
\end{equation}
which remains polynomial and significantly lower than the factorial complexity of exhaustive or tree search.
\begin{figure}[t]
    \centering
    \includegraphics[width=0.7\linewidth]{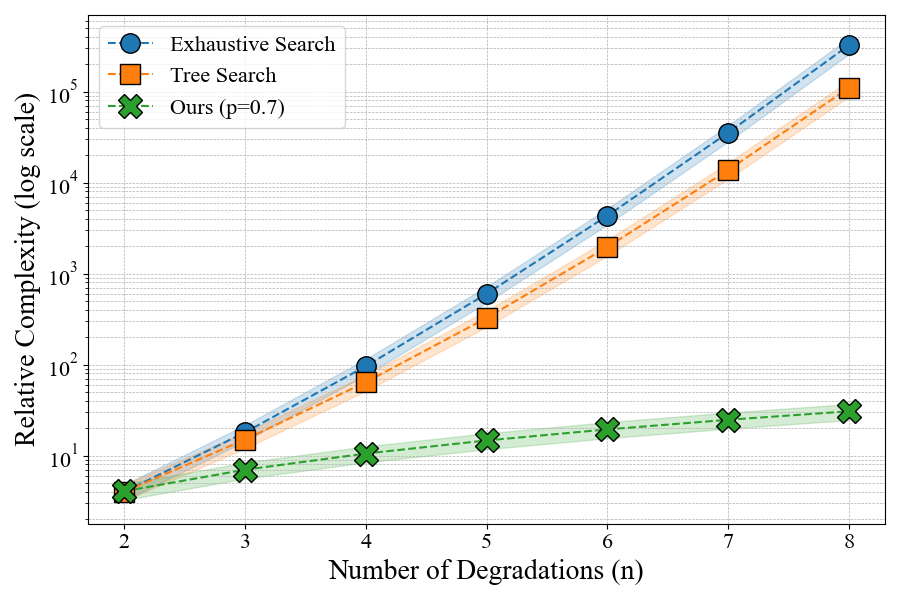}
    \caption{Time complexity comparison between three routing algorithms.}
    \label{fig:timecomplex}
\end{figure}

Fig.~\ref{fig:timecomplex} visualizes the time complexity comparison among three routing algorithms, indicating that our routing strategy achieves significantly lower complexity, especially as the number of degradations increases.

\begin{figure}[t]
    \centering
    \includegraphics[width=0.7\linewidth]{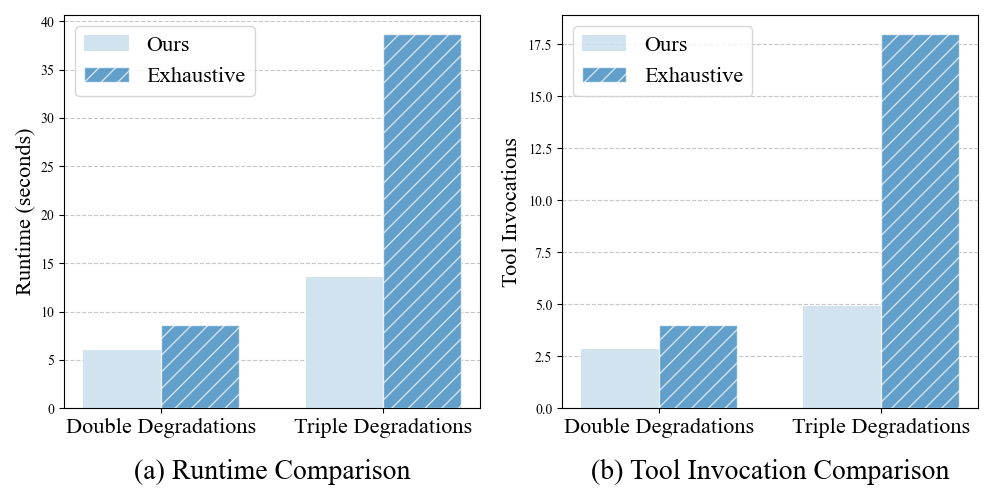}
    \caption{Additional experiments in runtime and tool invocation.}
    \label{fig:runtime}
\end{figure}

\textbf{Runtime and Tool Invocation Experiments.}
 We conduct additional experiments comparing our routing agent strategy with exhaustive search to further explore the routing efficiency of our approach. We fix a tool for each degradation removal and compared the time complexity of our routing strategy and the exhaustive search. The runtime per frame and tool invocations are reported in Fig.~\ref{fig:runtime}. It can be observed that, compared with exhaustive search, our method achieves a 66\% reduction in runtime on triple degradations.

\textbf{VQA-driven Optimization Baseline.} We augment BasicVSR++~\cite{chan2022generalization} with a differentiable VQA loss derived from our quality assessment agent. Specifically, in addition to the pixel-level Charbonnier reconstruction loss $\mathcal{L}_{\text{pixel}}$, we introduce a VQA loss that aligns the predicted perceptual quality of the restored video with that of the ground-truth video:
\[
\mathcal{L}_{\text{vqa}} = \frac{1}{B}\sum_{b=1}^{B} \big| \Phi(\hat V^{(b)}) - \Phi(V_{\text{gt}}^{(b)}) \big|,
\]
where $\Phi$ denotes the frozen VQA network, $B$ is the batch size, and $b \in \{1,\dots,B\}$ indexes the training samples.
During training, gradients are back\mbox{-}propagated only to the restoration backbone, while the VQA model remains fixed.
The overall objective is then defined as
\[
\mathcal{L}_{\text{total}}=\lambda_{\text{pixel}}\mathcal{L}_{\text{pixel}}+\lambda_{\text{vqa}}\mathcal{L}_{\text{vqa}},
\]
with $\lambda_{\text{vqa}}=10^{-4}$. We follow the standard BasicVSR++ fine-tuning protocol and schedule, using MoA-VD as training data.

Table~\ref{table6} shows that in both double- and triple-degradation settings, the VQA-driven baseline increases perceptual quality but reduces pixel-level fidelity (e.g., under double degradation: $\Delta$MUSIQ $+2.77$, $\Delta$PSNR $-0.33\,\mathrm{dB}$), and it remains clearly inferior to MoA-VR overall. Notably, the VQA-driven baseline lacks explicit degradation disentanglement and routing, which limits its ability to robustly handle compound degradations.Notably, the VQA-driven baseline lacks explicit degradation disentanglement and routing, which limits its ability to robustly handle compound degradations.
\setcounter{table}{8}
\renewcommand{\arraystretch}{1.2} 
\begin{table}[t]
\caption{
Comparison of MoA-VR with All-in-One methods for multi-degraded video restoration. We
highlight \textbf{best} in \textbf{bold}.
}
\label{table6}
\centering
\resizebox{\linewidth}{!}{%
\setlength{\tabcolsep}{3pt} %
\begin{tabular}{@{}lcccccccc}
\toprule
             & \multicolumn{4}{c}{\mytext{Double Degradation}}&\multicolumn{4}{c}{\mytext{Triple Degradation}}\\ \cmidrule(l){2-5} \cmidrule(l){6-9} 
             & PSNR \upcolor{$\uparrow$}  & SSIM \upcolor{$\uparrow$}   & LPIPS \upcolor{$\downarrow$}  & MUSIQ\upcolor{$\uparrow$}& PSNR \upcolor{$\uparrow$}  & SSIM \upcolor{$\uparrow$}   & LPIPS \upcolor{$\downarrow$}  & MUSIQ\upcolor{$\uparrow$}\\
BasicVSR++~\cite{chan2022generalization}& 20.45& 0.6049& 0.5146& 25.65& 17.09& 0.5026& 0.5579& 24.65\\
\rowcolor{gray!20} BasicVSR++ + VQA loss& 20.12& 0.5981& 0.4013& 28.42& 16.95& 0.4897& 0.4126& 26.31\\
\rowcolor{gray!20} \quad $\Delta$ $\uparrow$ (vs. BasicVSR++) & -0.33 & -0.0068 & -0.1133 & +2.77 & -0.14 & -0.0129 & -0.1453 & +1.66\\
\mytext{MoA-VR-Ours}& \textbf{23.47}& \textbf{0.6852}& \textbf{0.3386}& \textbf{36.14}& \textbf{22.94}& \textbf{0.6497}& \textbf{0.3074}& \textbf{30.43}\\
\bottomrule
\end{tabular}%
}
\end{table}

\section{Conclusion}
\label{con}
We introduce MoA-VR, a novel framework for all-in-one video restoration especially in complex mixed degradation scenarios. This system is built upon a mixture-of-agent architecture, comprising three collaborative agents: degradation identification agent, routing and restoration agent, and restoration quality assessment agent. These agents work together in a closed-loop process to emulate the reasoning of human experts. Specifically, we utilize a VLM to develop the degradation identification agent and evaluate its predictive performance. Additionally, we incorporate an agent driven by an LLM that autonomously adapts routing strategies by observing the effect of degradation removal. To assess the quality of the output, we create a restoration video quality (Res-VQ) dataset and develop a VQA model focused on restoration tasks. Extensive experiments validate that MoA-VR successfully handles diverse and compound degradation issues, consistently outperforming existing benchmarks in both objective metrics and perceptual quality. These findings emphasize the potential of integrating multimodal intelligence and modular reasoning into general-purpose video restoration systems.

\bibliographystyle{IEEEtran}
\bibliography{main}

\end{document}